%% file: acl_latex.tex
\documentclass[11pt]{article}

\usepackage[preprint]{acl}

\usepackage{times}
\usepackage{latexsym}

\usepackage[T1]{fontenc}

\usepackage[utf8]{inputenc}

\usepackage{microtype}

\usepackage{inconsolata}

\usepackage{graphicx}

\usepackage{booktabs}
\usepackage{amssymb}
\usepackage{pifont} 

\usepackage{makecell}
\usepackage{float}
\usepackage{hyperref}
\usepackage{url}
\PassOptionsToPackage{prologue,dvipsnames, x11names, table}{xcolor}
\usepackage[dvipsnames]{xcolor}
\usepackage[x11names]{xcolor}
\usepackage[table]{xcolor}
\usepackage[most]{tcolorbox}
\usepackage{colortbl}
\usepackage{bbding}
\usepackage{pifont}
\usepackage{booktabs}
\usepackage{capt-of}
\usepackage{subfigure}
\usepackage{longtable}
\usepackage{titletoc}
\usepackage{enumitem}
\usepackage{multirow}
\usepackage{wrapfig}
\usepackage{graphicx}
\usepackage{arydshln}
\usepackage{array} 
\tcbuselibrary{breakable,skins,listings}
\usepackage{listings}
\newcolumntype{C}[1]{>{\centering\arraybackslash}p{#1}}

\usepackage{CJKutf8}

\title{Aligned Multi-View Scripts for Universal Chart-to-Code Generation}

\author{
 \textbf{Zhihan Zhang},
 \textbf{Lizi Liao}
\\
 School of Computing and Information Systems, \\Singapore Management University
 \\
 \texttt{zhihanzhang.2024@phdcs.smu.edu.sg, lzliao@smu.edu.sg}
}

\begin{document}
\maketitle
\begin{abstract}
Chart-to-code generation converts a chart image into an executable plotting script, enabling faithful reproduction and editable visualizations. Existing methods are largely Python-centric, limiting practical use and overlooking a critical source of supervision: the same chart can be expressed by semantically equivalent scripts in different plotting languages. To fill this gap, we introduce \textbf{Chart2NCode}, a dataset of 176K charts paired with aligned scripts in Python, R, and LaTeX that render visually equivalent outputs, constructed via a metadata-to-template pipeline with rendering verification and human quality checks. Building on a LLaVA-style architecture, we further propose \textbf{CharLuMA}, a parameter-efficient adaptation module that augments the multimodal projector with a language-conditioned mixture of low-rank subspaces, allowing the model to share core chart understanding while specializing code generation to the target language through lightweight routing. Extensive experiments show consistent gains in executability and visual fidelity across all languages, outperforming strong open-source baselines and remaining competitive with proprietary systems. Further analyses reveal that balanced multi-language supervision benefits all languages and that the adapter allocates a compact shared core plus language-specific capacity
\footnote{Codes and data are available at \url{https://github.com/Zhihan72/CharLuMA}.}.
\end{abstract}

\input{sections/1_introduction}
\input{sections/2_related_work}
\input{sections/3_dataset}
\input{sections/4_model}
\input{sections/5_experiment}
\input{sections/6_analysis}

\section{Conclusion}

We leverage the visual semantic equivalence of scripts in different plotting languages to drive universal chart-to-code generation.
We introduce Chart2NCode, a pioneering dataset of 176K visually aligned Chart–Python–R–LaTeX quadruples, and CharLuMA that realizes the multi-view supervision via language-conditioned routing over low-rank subspaces. 
Extensive experiments show that balanced multi-language supervision from aligned scripts provides complementary training signals that improve executability and visual fidelity over strong baselines. These contributions pave the way toward universal chart-to-code systems that reflect the diverse software ecosystems in practice.

\section*{Limitations}
While this study offers a comprehensive analysis, we acknowledge specific limitations that merit future investigation. First, constrained by computational resources, we limited the CharLuMA architecture to 1.3B and 6.7B parameters. Although these scales demonstrate robust performance, scaling to larger backbones may yield further improvements in reasoning and generation quality. Second, the model remains constrained by the fixed input resolution of the visual encoder. While our choice of SigLIP represents a significant improvement over baselines like CLIP-Large, resolution bottlenecks may still limit performance on information-dense charts. Future work will focus on integrating high-resolution visual adapters to better handle these visually complex scenarios.

\section*{Acknowledgments}
This research was supported by the National Research
Foundation, Singapore under its National Large Language Models Funding Initiative (AISG Award
No. AISG-NMLP-2024-002), and by the Ministry of Education, Singapore, under its AcRF Tier 2
Funding (Proposal ID: T2EP20123-0052). Any opinions, findings, conclusions, or recommendations
expressed in this material are those of the author(s) and do not reflect the views of the National
Research Foundation or the Ministry of Education, Singapore.

\bibliography{custom}

\appendix

\input{sections/appendix}

\end{document}

%% file: sections/1_introduction.tex
\section{Introduction}

Charts serve as a compact and prevalent medium for communicating quantitative evidence in scientific literature, but they are frequently disseminated as static images, which impedes reproduction, editing, and reuse. Chart-to-code generation \citep{shi2024chartmimicevaluatinglmmscrossmodal} bridges this gap by translating a chart image into an executable plotting script that faithfully reconstructs both the underlying data encodings and the visual design attributes. While recent multimodal large language models (MLLMs) have demonstrated strong capabilities in general vision–language tasks \citep{10656299, Lu2023MathVistaEM, zhang-etal-2025-xfinbench}, chart-to-code generation requires a substantially higher degree of precision: minor discrepancies in extracted data, axis specifications, or stylistic parameters can result in compilation failures or visually misleading outputs. 
Furthermore, the field remains constrained by a restrictive Python/matplotlib-centric bias \citep{wu-etal-2025-plot2code, zhao-etal-2025-chartcoder, zhihan2025chart2code}, neglecting the diverse ecosystem of plotting tools—such as R (ggplot2) and LaTeX (TikZ)—that serve as publication standards in many scientific disciplines \citep{kaggle-survey}.

\begin{figure}[t]
    \centering
    \includegraphics[width=\linewidth]{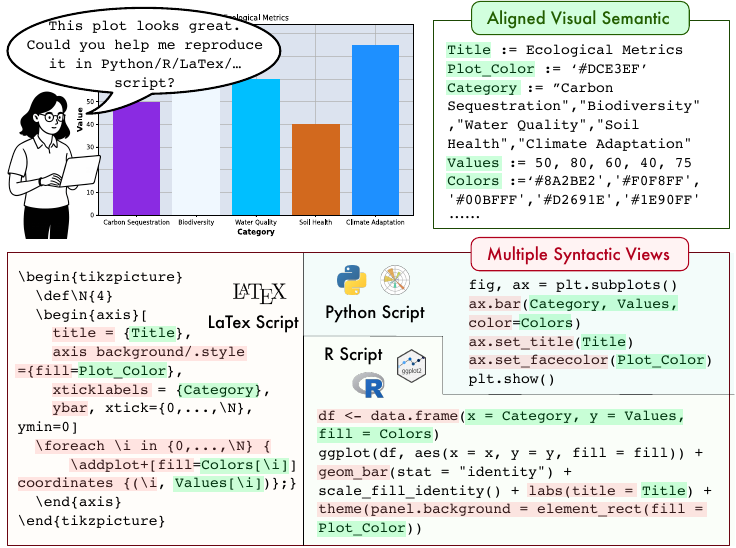}
    \caption{Illustration of aligned multi-view scripts for chart-to-code generation task.}
    \label{fig:overview}
    \vspace{-0.2em}
\end{figure}

Beyond real-world applicability, the Python-centric paradigm overlooks a critical learning signal: the alignment of visual semantics across different plotting languages, as shown in Figure~\ref{fig:overview}. While surface syntax diverges, scripts in multiple languages share a common latent visual semantic, encoding identical data tables, stylistic designs, and layout constraints. We frame these distinct scripts as complementary ``views'' of the same underlying chart semantics. We use the alignment across languages to provide multi-target supervision for chart-to-code learning, where the same chart can be realized in multiple plotting syntaxes.

Realizing this multi-view idea necessitates a resource currently absent from the literature: aligned chart–code pairs across multiple languages that are visually isomorphic. 
The predominantly monolingual nature of existing datasets restricts models to a single target syntax, precluding the study of cross-language supervision \citep{shi2024chartmimicevaluatinglmmscrossmodal,zhao-etal-2025-chartcoder, wu-etal-2025-plot2code, niu-etal-2025-chart2code53}. 
To bridge this gap, we introduce \textbf{Chart2NCode}, a pioneering dataset of 176K chart images paired with aligned scripts in Python, R, and LaTeX. 
The dataset is constructed via an automated pipeline that synthesizes language-agnostic metadata into language-specific templates, ensuring high fidelity through rendering verification and human quality checking.
This dataset supports both training and evaluation of multi-language chart-to-code models under consistent supervision.

A second challenge lies in the modeling: developing separate experts for each language is inefficient and discards the shared structure, while straightforward multi-language training can suffer from interference and uneven specialization.
We propose \textbf{CharLuMA}, a parameter-efficient adaptation approach that preserves shared chart understanding while enabling language-specific code realization. 
Building on the LLaVA architecture \citep{liu2023visual}, CharLuMA augments the multimodal projector with a language-conditioned mixture of low-rank subspaces (Figure~\ref{fig:model_archi}).
This module operates via a lightweight routing mechanism, which dynamically selects and combines a small subset of subspaces conditioned on the target language and visual features. 
This design enables the model to reuse a compact shared core while adapting its latent representations to specific syntactic conventions, offering an efficient alternative to the redundancy of independent experts or the interference of joint training.
Experimental results demonstrate that balanced multi-language alignment yields consistent improvements across all languages and that the adapter allocates a compact shared core plus language-specific capacity.

To sum up, our contributions are three-fold:
\begin{itemize}[leftmargin=*,topsep=-1pt]
\setlength{\parskip}{1pt}
\setlength{\itemsep}{0pt plus 1pt}
    \item We formulate chart-to-code generation in a multi-language setting and propose CharLuMA that realizes multi-view script alignment via language-guided routing over low-rank subspaces.
    \item We present Chart2NCode, a dataset of 176K visually aligned chart–Python–R–LaTeX quadruples that enables the first systematic study of universal chart-to-code generation beyond Python.
    \item Extensive experiments validate our approach against open-source baselines and confirm that diverse, balanced multi-language supervision synergistically benefits all languages.
\end{itemize}

%% file: sections/2_related_work.tex
\section{Related Work}
\label{sec:related_work}

\begin{figure*}[t]
    \centering
    \includegraphics[width=\linewidth]{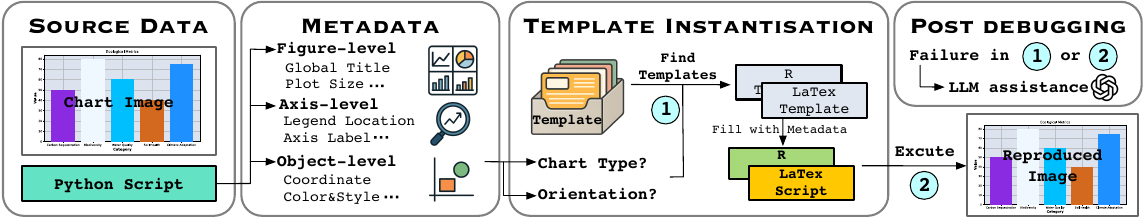}
    \caption{Overview of the automatic annotation pipeline of Chart2NCode.}
    \label{fig:data_annot}
    \vspace{-0.2em}
\end{figure*}

\noindent \textbf{Multimodal large language models} employ multimodal projectors to bridge vision encoders with large language models, enabling reasoning across modalities. 
Existing works explores diverse strategies to optimize visual perception, ranging from compressing visual tokens \citep{Li2022BLIPBL, Bai2023QwenVLAF,wenbo2024BLIVA} to fusing multiple vision encoders \citep{NEURIPS2024_9ee3a664, lin2023sphinxjointmixingweights}.
LLaVA \citep{liu2023visual,10655294} demonstrates that a simple MLP projector can effectively align modalities without discarding visual information.
Some recent works employ sparsely gated MoE projectors \citep{10.5555/3737916.3742086, xu2025chartmoe}, which parallelize MLPs as experts at the cost of substantial parameter growth.

\noindent \textbf{Chart-to-code generation} task requires models to translate chart images into executable plotting scripts, challenging MLLMs with demands in visual understanding, code generation, and cross-modal reasoning. Prior efforts have primarily focused on chart-to-Python generation, spanning dataset construction \citep{shi2024chartmimicevaluatinglmmscrossmodal, zhao-etal-2025-chartcoder, niu-etal-2025-chart2code53}, multi-agent framework \citep{yang-etal-2024-matplotagent, 10.1145/3701716.3716888} , and preference learning method \citep{zhihan2025chart2code}.
Other studies untilize chart-to-code generation for aligning multimodal projectors \citep{xu2025chartmoe} or constructing question answering datasets \citep{zhang-etal-2024-multimodal, he2025distillvisualchartreasoning}. 
These efforts remain restricted to single-language settings, which limits practical applicability and overlooks the learning signals in cross-language alignment.

\noindent \textbf{Multi-view representation learning} aims to construct comprehensive representations by integrating complementary information from multiple distinct views. It often enforces consistency constraints, extracting robust features that remain invariant across these views \citep{tian2020contrastive, bachman2019amdim}. This principle has proven effective in both natural language processing \citep{lample2019cross, conneau2020unsupervised} and code generation \citep{lachaux2020unsupervised, guo2022unixcoder}, where models align diverse surface syntaxes into a shared semantic space. We extend this paradigm to scientific visualization, treating Python, R, and LaTeX scripts as complementary views of a single chart.

%% file: sections/3_dataset.tex
\section{The Chart2NCode Dataset}
\label{sec:dataset}

We present Chart2NCode, the first large-scale dataset that aligns chart-code pairs across multiple programming languages. With 176K Chart-Python-R-LaTeX quadruples, Chart2NCode establishes a comprehensive resource for developing and evaluating multi-language chart-to-code models.

\subsection{Automatic Annotation}

We construct multi-language plotting scripts via an automatic annotation pipeline as shown in Figure~\ref{fig:data_annot}. We first collect single-language source data from publicly available and open-source repositories. Specifically, we utilize Python scripts from ChartCoder \citep{zhao-etal-2025-chartcoder}, a chart-specific subset of LaTeX scripts from DaTikZ \citep{belouadi2024automatikz}, and 40K newly curated R scripts from online communities, ensuring compliance with their respective open licenses and terms of use.

\noindent \textbf{Metadata Extraction.}
We extract language-agnostic metadata from single-language plotting scripts at the figure, axis, and object levels. 
The figure level captures global attributes that determine the overall layout and presentation of the chart. The axis level records structural elements that define the coordinate system and its descriptive properties. The object level encodes graphical primitives together with their visual styles, ensuring precise representation of chart content. 
Metadata is obtained from plotting objects for Python (\texttt{matplotlib.axes}) and R (\texttt{ggplot\_build()}), while LaTeX scripts are processed via regular-expression parsing.
Collectively, these layers yield a comprehensive description of each chart, enabling faithful reconstruction using different languages in the following steps.

\noindent \textbf{Template Instantiation.} 
We synthesize multi-language scripts by identifying object-level metadata patterns to retrieve and instantiate language-specific templates. 
For instance, a horizontal bar chart is characterized at the object level by rectangles of equal height and varying width, which are organized into a data table and matched to the corresponding templates in different languages. Our library comprises 202 human-curated templates spanning over 30 chart subtypes in Python, R, and LaTeX, derived from systematic observations of the source data. Once the appropriate template is identified, it is instantiated with structured metadata such as titles, axis ticks, and data values. We also add an attribute-mapping process during instantiation to maintain cross-language consistency, such as mapping the \texttt{bold} font style in Python to the \texttt{bfseries} directive in LaTeX.

\noindent \textbf{Post Debugging.} 
In situations where template identifying is unsuccessful or script execution errors occur, we incorporate an LLM-assisted debugging module powered by GPT-4o \citep{gpt-4o-website}. If no suitable template exists, the module translates the available single-language script into the target languages; if an instantiated template fails, it applies error correction to restore executability. Scripts that remain invalid or produce deprecated figures are discarded to maintain dataset quality. 

We detail the source data acquisition, annotation pipeline, and illustrative examples in Appendix~\ref{sec:data_acquisition}, Appendix~\ref{sec:annot_pipeline}, and Appendix~\ref{sec:case_study}, respectively.

\subsection{Human Quality Checking}
\label{sec:quality_check}

We conduct human evaluation to assess the cross-language fidelity of Chart2NCode using a random sample of 1,000 quadruples. We recruited three independent annotators from the university campus, requiring demonstrated proficiency in data visualization across all target languages. Annotators evaluated the samples across four dimensions on a 1–5 scale. The proportion of examples achieving an average score $\geq$ 4 demonstrates high quality across all dimensions: structural fidelity (98.2\%), data integrity (95.2\%), semantic consistency (97.4\%), and stylistic coherence (95.8\%).
The inter-annotator agreement yields a Krippendorff's $\alpha$ \citep{krippendorff2011computing} of 0.81, indicating robust evaluation reliability. Further details and results are in Appendix~\ref{sec:data_quality_check}.

\subsection{Data Statistics}

Chart2NCode comprises 176k quadruples consisting of a chart image and aligned scripts in Python, R, and LaTeX, with 14.7\% refined via LLM-assisted debugging.
In terms of diversity, the dataset covers a wide spectrum of 20 distinct chart types, ranging from 18 regular types to 2 advanced categories featuring composite layouts and multiple coordinate systems.
For benchmarking purpose, we construct a test set of 1,000 randomly sampled examples that achieve average scores of at least 4 across all quality aspects in Section \ref{sec:quality_check}.
Using the Llama 3 tokenizer \citep{llama-3-website}, we observe average token counts of 384.1 for Python, 591.8 for R, and 637.1 for LaTeX, with standard deviations of 189.7, 242.0, and 247.1, respectively.
Detailed statistics are provided in Appendix \ref{sec:detailed_data_stat}.

%% file: sections/4_model.tex
\section{The CharLuMA Model}
\label{sec:model}

We propose CharLuMA, a chart-to-code MLLM that extends a LLaVA-style architecture with a novel low-rank subspace adapter for efficient multi-language adaptation. The model is optimized via a progressive training strategy that combines alignment pretraining with instruction tuning.

\subsection{Architecture}

\begin{figure}[t]
    \centering
    \includegraphics[width=\linewidth]{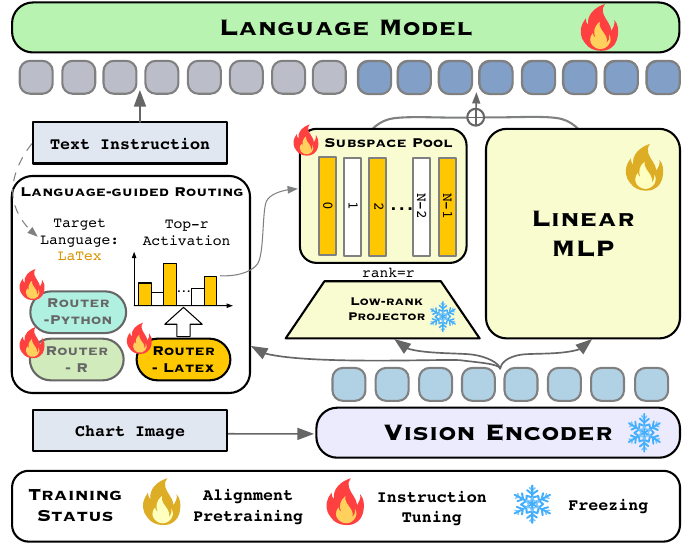}
    \caption{Overview of CharLuMA. The adapter employs language-conditioned routing to dynamically compose low-rank subspaces, exemplified here for a LaTeX target. The training strategy comprises alignment pretraining followed by instruction tuning.}
    \label{fig:model_archi}
    \vspace{-0.2em}
\end{figure}

CharLuMA is composed of a vision encoder and a LLM backbone, connected through a two-layer MLP projector augmented with a novel low-rank subspace adapter (see Figure~\ref{fig:model_archi}). The adapter is governed by a language-guided routing policy that dynamically selects subspace experts based on both the chart’s image features and the target language token, enabling language-specific specialization while maintaining shared visual understanding.

\noindent \textbf{Vision Encoder.} 
We adopt SigLIP \citep{10377550} as the vision encoder, configured with an input resolution of $384\times384$. Pretrained on millions of image–text pairs, it provides strong priors for extracting semantically meaningful visual features. Formally, given a chart input $\mathbf{X}_{v}$, the vision encoder $g^v(\cdot)$ generates its corresponding representation $\mathbf{Z}_{v}$, \textit{i.e.} $ \mathbf{Z}_{v} = g^v( \mathbf{X}_{v})$.

\noindent \textbf{Multimodal Projector.}
The standard multimodal projector in LLaVA-style architectures \citep{liu2023visual} is a two-layer MLP block $\mathbf{W}$ that performs a one-to-one transformation, mapping visual features $\mathbf{Z}_{v}$ into the embedding space of the LLM backbone. The resulting output, $\mathbf{H}_{\text{base}} = \mathbf{W} \mathbf{Z}_{v}$, serves as a shared base representation across languages.

To enable efficient language adaptation while preserving visual understanding, we augment this linear MLP block with a low-rank subspace adapter \citep{ding-etal-2025-sulora, wu-etal-2024-mixture-subspaces}. 
The adapter consists of three components: a low-rank projector $\mathbf{A}$, a subspace pool $\{b_i\}_{i=1}^N$, and language-specific routers $\mathbf{W}^l$ ($l \in \{\text{Python}, \text{R}, \text{LaTeX}\}$) \citep{10377734}. 
Given the visual features $\mathbf{Z}_{v}$, the projector $\mathbf{A}$ maps them into a compact rank-$r$ representation ($r < N$). The router then determines which subspaces to activate for the target language $l$, as specified in the text instruction.
Concretely, the router $\mathbf{W}^l$ applies a language-specific transformation to the mean-pooled visual feature $\overline{\mathbf{Z}}_{v}$, yielding a probability distribution over the subspace pool. 
The top-$r$ subspaces are then selected,
$y^l=\text{top}_\text{r}(\text{softmax}( \mathbf{W}^l \overline{\mathbf{Z}}_{v}))$ 
where $y^l$ denotes their indices, and concatenated to form the matrix $\mathbf{B} = \text{concat}_{i\in y^l} b_i$.
The reconstruction matrix $\mathbf{B}$ is combined with the low-rank projector $\mathbf{A}$ to map the visual features into the LLM embedding space, yielding an language-adaptable representation. The final visual tokens injected into the LLM consists of visual tokens that merge the base and language-adaptable representations:
\begin{equation*}
\setlength\abovedisplayskip{3pt}
\setlength\belowdisplayskip{3pt}
 \mathbf{H}_{v} = \mathbf{H}_{\text{base}} + \mathbf{H}_{\text{adapt}} = \mathbf{W} \mathbf{Z}_{v} + \mathbf{A} \mathbf{B} \mathbf{Z}_{v}.
\end{equation*}

\noindent \textbf{Large Language Model.}
We use DeepSeek-Coder \citep{guo2024deepseekcoderlargelanguagemodel} as the LLM backbone, with 1.3B and 6.7B variants named CharLuMA-1.3B and CharLuMA-6.7B. The visual tokens $\mathbf{H}_{v}$ produced by the multimodal projector are concatenated with the text tokens $\mathbf{H}_{t}$ to construct the input sequence for the LLM $g^{L}(\cdot)$. The final output is then obtained as $ g^{L}(\mathbf{H}_{v}; \mathbf{H}_{t})$.

\subsection{Training Strategy}
\label{sec:training_strategy}

\noindent \textbf{Alignment Pretraining.} 
We initialize the multimodal projector by pretraining the linear MLP block $\mathbf{W}$ on ChartMoE-Align \citep{xu2025chartmoe}, a dataset comprising 900k Chart–JSON pairs that encode structural representations. The vision encoder and LLM backbone remain frozen during this stage, ensuring that $\mathbf{W}$ learns to align visual features of charts with textual schema representations without altering pretrained components \citep{10.1007/978-3-031-70533-5_26}.

\noindent \textbf{Instruction Tuning.} 
We augment the multimodal projector with the proposed low-rank subspace adapter and finetune the model on Chart2NCode. We first warm up the language-specific routers $\mathbf{W}^l$ and the subspace pool $\{b_i\}_{i=1}^N$ over fixed steps, while keeping the MLP block, vision encoder, and LLM backbone frozen. The low-rank projector $\mathbf{A}$ is randomly initialized and kept frozen throughout training, ensuring that adaptation capacity is directed toward language-specific diversities rather than redundantly modeling visual commonalities \citep{ding-etal-2025-sulora, 10.5555/3737916.3738220}. We then unfreeze the LLM backbone and continue training jointly with the routers and subspace pool, while keeping the MLP block, vision encoder, and $\mathbf{A}$ frozen. This progressive protocol stabilizes routing and subspace specialization in the early phase, and subsequently enables the LLM to effectively leverage language-adaptive visual tokens.

\begin{table*}[t]
    \belowrulesep=0pt
    \aboverulesep=0pt
    \renewcommand{\arraystretch}{1.15}
    \setlength{\belowcaptionskip}{0.1cm}
    \centering
    \setlength\tabcolsep{4pt}
    \scalebox{0.78}{
    \begin{tabular}{p{3.0cm}|p{0.6cm}<{\centering}p{0.6cm}<{\centering}p{0.6cm}<{\centering}p{0.9cm}<{\centering}|p{0.6cm}<{\centering}p{0.6cm}<{\centering}p{0.6cm}<{\centering}p{0.9cm}<{\centering}|p{0.6cm}<{\centering}p{0.6cm}<{\centering}p{0.6cm}<{\centering}p{0.9cm}<{\centering}|p{0.6cm}<{\centering}p{0.6cm}<{\centering}p{0.9cm}<{\centering}|p{0.6cm}<{\centering}p{0.6cm}<{\centering}p{0.6cm}<{\centering}}
    \toprule
    \multirow{3}{*}{Models} & \multicolumn{4}{c}{\textbf{ChartMimic}}
                            & \multicolumn{4}{c}{\textbf{Plot2Code}}
                            & \multicolumn{10}{c}{\textbf{Chart2NCode}} \\
    \cmidrule(r){2-5} \cmidrule(r){6-9} \cmidrule(r){10-19}
     & \multicolumn{4}{c}{Chart2Python}
     & \multicolumn{4}{c}{Chart2Python}
     & \multicolumn{4}{c}{Chart2Python}
     & \multicolumn{3}{c}{Chart2R}
     & \multicolumn{3}{c}{Chart2LaTeX} \\
    \cmidrule(r){2-5} \cmidrule(r){6-9} \cmidrule(r){10-13} \cmidrule(r){14-16}  \cmidrule(r){17-19}
     & ER & DS & MJ & \multicolumn{1}{c}{F1} & ER & DS & MJ & \multicolumn{1}{c}{F1} & ER & DS & MJ & \multicolumn{1}{c}{F1} & ER & \multicolumn{1}{c}{DS} & MJ & ER & DS & MJ \\
    \midrule
    & \multicolumn{18}{c}{\textit{Propriety Multimodal Large Language Models}} \\
    \midrule
    GPT-5-mini & 86.8 & 86.9 & 78.2 & 71.5 & 93.2 & 85.9 & 72.7 & 72.8 & 85.2 & 89.0 & 80.0 & 67.5 & 90.3 & 82.5 & 81.2 & 49.7 & 75.2 & 41.1 \\
    GPT-4o-mini & 89.0 & 77.5 & 74.8 & 70.2 & 90.2 & 77.8 & 63.8 & 67.0 & 94.8 & 81.2 & 79.8 & 74.5 & 89.5 & 75.4 & 70.3 & 94.7 & 61.2 & 70.4 \\
    GPT-4o & 93.2 & 83.5 & 83.5 & 79.0 & 92.4 & 83.6 & 76.7 & 75.4 & 98.5 & 85.0 & 87.4 & 80.9 & 94.5 & 78.8 & 78.3 & 88.4 & 72.4 & 69.8 \\
    Claude-Haiku-3.5 & 88.0 & 76.2 & 73.5 & 65.7 & 87.1 & 72.8 & 60.6 & 56.8 & 91.3 & 81.6 & 76.7 & 68.8 & 93.0 & 76.2 & 73.9 & 78.2 & 57.3 & 55.3 \\
    Claude-Sonnet-4 & 96.2 & 83.3 & 86.4 & 81.5 & 95.5 & 81.2 & 81.0 & 76.8 & 98.3 & 86.8 & 88.0 & 81.4 & 93.9 & 82.0 & 83.1 & 92.7 & 76.0 & 72.2 \\
    \midrule
    & \multicolumn{18}{c}{\textit{Open-source Multimodal Large Language Models}} \\
    \midrule
    Qwen3-VL-2B & 56.3 & 68.4 & 39.7 & 39.2 & 68.9 & 64.2 & 41.2 & 50.1 & 74.0 & 78.0 & 59.6 & 61.0 & 56.5 & 52.4 & 42.0 & 56.0 & 60.8 & 37.4 \\
    Qwen3-VL-4B & 72.5 & 71.9 & 58.4 & 55.2 & 77.3 & 66.4 & 54.4 & 55.4 & 87.6 & 83.2 & 77.2 & 76.1 & 75.4 & 66.4 & 60.9 & 62.4 & 68.6 & 45.2 \\
    Qwen3-VL-8B & 78.7 & \underline{72.5} & 65.2 & 61.9 & 78.8 & \underline{68.1} & 57.3 & \underline{56.9} & 91.1 & 83.7 & 80.8 & \underline{80.6} & 73.6 & 72.7 & 57.2 & 77.3 & 66.8 & 57.1 \\
    InternVL3.5-2B & 48.5 & 65.6 & 32.5 & 31.3 & 61.4 & 55.7 & 34.2 & 44.2 & 69.8 & 76.1 & 53.2 & 53.1 & 61.8 & 53.4 & 44.9 & 9.6 & 52.6 & 4.7 \\
    InternVL3.5-4B & 62.8 & 69.5 & 44.7 & 43.0 & 62.1 & 58.8 & 38.0 & 42.7 & 77.9 & 78.4 & 63.4 & 63.0 & 66.8 & 56.4 & 51.5 & 25.7 & 55.5 & 14.7 \\
    InternVL3.5-8B & 71.2 & 71.0 & 52.1 & 48.9 & 74.2 & 61.0 & 47.2 & 49.1 & 82.5 & 79.6 & 67.5 & 67.0 & 67.0 & 67.6 & 48.2 & 81.1 & 57.1 & 53.3 \\
    DeepSeek-VL2-3B & 50.2 & 69.4 & 35.0 & 35.0 & 72.5 & 59.8 & 45.4 & 44.2 & 72.0 & 80.2 & 58.7 & 58.0 & 23.0 & 56.9 & 16.4 & 3.9 & 42.7 & 1.9 \\
    Phi-3.5-vision-4B & 66.7 & 44.1 & 41.0 & 38.6 & 72.7 & 63.8 & 43.1 & 42.6 & 68.8 & 53.3 & 56.1 & 34.2 & 47.0 & 52.5 & 33.5 & 7.9 & 42.9 & 5.1 \\
    LLaVA-v1.6-7B & 62.0 & 55.9 & 25.7 & 25.9 & 69.7 & 49.8 & 26.5 & 32.5 & 71.0 & 65.7 & 38.0 & 39.5 & 58.5 & 50.9 & 37.9 & 77.3 & 42.4 & 42.2 \\
    ChartLlama-13B & 57.5 & 44.9 & 28.1 & 24.8 & 81.8 & 50.1 & 18.9 & 22.4 & 65.3 & 46.0 & 14.8 & 16.2 & 13.0 & 44.8 & 6.2 & 21.7 & 32.5 & 10.2 \\
    ChartMoE-8B & 52.7 & 56.9 & 22.9 & 25.3 & 70.5 & 58.9 & 37.6 & 26.9 & 69.5 & 64.2 & 40.2 & 35.4 & 39.3 & 52.9 & 25.5 & 17.1 & 27.9 & 11.1 \\
    ChartCoder-7B & \underline{91.4} & 69.2 & \underline{74.0} & \underline{72.5} & \underline{87.9} & 65.7 & \underline{58.2} & 56.6 & \underline{96.2} & 48.1 & \underline{86.4} & 56.1 & - & - & - & 17.9 & 39.1 & 10.6 \\
    CharLuMA-1.3B & 83.0 & 71.8 & 64.8 & 62.5 & 83.3 & 64.3 & 50.6 & 47.2 & 94.4 & \underline{86.5} & 78.4 & 76.9 & \underline{94.5} & \underline{78.9} & \underline{73.3} & \underline{84.5} & \underline{71.3} & \underline{65.1} \\
    CharLuMA-6.7B & \textbf{92.3} & \textbf{77.4} & \textbf{75.1} & \textbf{73.3} & \textbf{96.2} & \textbf{68.3} & \textbf{62.2} & \textbf{60.5} & \textbf{98.0} & \textbf{88.7} & \textbf{88.1} & \textbf{83.5} & \textbf{96.5} & \textbf{81.8} & \textbf{80.9} & \textbf{89.0} & \textbf{72.5} & \textbf{74.2} \\
    \bottomrule
    \end{tabular}
    }
    \caption{Performance on ChartMimic, Plot2Code, and Chart2NCode test set. $\text{ER}\Uparrow$ denotes execution rate, $\text{DS}\Uparrow$ denotes the image-similarity score DreamSim, $\text{MJ}\Uparrow$ denotes the MLLM-as-Judge score, and $\text{F1}\Uparrow$ denotes the heuristic F1 score for Python scripts. A ``-'' indicates that no executable script is generated.} 
    \label{table:experimental_result}
\end{table*}

%% file: sections/5_experiment.tex
\section{Experiment}
\label{sec:experiment}

We validate the efficacy of CharLuMA through extensive experiments across diverse benchmarks, establishing consistent gains over strong baselines in multi-language chart-to-code generation.

\subsection{Implementation Details}
\label{sec:implementation_details}

During alignment pretraining, we train the MLP block for 1 epoch on 900k Chart–JSON pairs from ChartMoE-Align \citep{xu2025chartmoe}, with a learning rate of 2e-4. 
During instruction tuning, we warm up the subspace pool and language-specific routers for 274 steps, and then continue with full fine-tuning of the LLM backbone for 1 epoch on the Chart2NCode training set, which contains 175k Chart–Python–R–LaTeX quadruples.
The learning rates are set to 2e-4 for the warm-up phase and 2e-5 for fine-tuning. 
We set the subspace size $N=32$ and the rank $r=16$. Detailed experimental settings are provided in Appendix \ref{sec:hyperparam_setting}.

\subsection{Evaluation Settings}

\noindent \textbf{Datasets.}
We evaluate CharLuMA and baselines on three chart-to-code datasets. The Chart2NCode test set provides 1,000 charts paired with scripts in Python, R, and LaTeX, enabling multi-language evaluation. ChartMimic \citep{shi2024chartmimicevaluatinglmmscrossmodal} testmini set comprises 600 charts with human-curated matplotlib scripts in Python, spanning 22 chart types. Plot2Code \citep{wu-etal-2025-plot2code} contains 132  high-quality matplotlib plots across 6 types.

\noindent \textbf{Evaluation Metrics.} 
We evaluate chart-to-code performance across two primary dimensions: executability and fidelity. 
Execution Rate (ER) measures the proportion of generated scripts that run successfully. 
To assess fidelity, we adopt DreamSim (DS) \citep{fu2023dreamsim}, a metric capturing perceptual similarity between generated and ground-truth images. 
Following the successful use of large foundation models for evaluation in computer vision \citep{shi2024chartmimicevaluatinglmmscrossmodal,zhao-etal-2025-chartcoder}, we further employ an MLLM-as-Judge approach (MJ) to assess visual alignment of reproduced charts. We employ GPT-4o as the judge following criteria in Figure~\ref{fig:prompt_gpt_scoring}, with reproducibility validated against open-source alternatives with high correlation results (see Appendix \ref{sec:mllm_judge}).
For Python scripts specifically, we report an averaged F1 score covering text, layout, type, and color attributes \citep{shi2024chartmimicevaluatinglmmscrossmodal}.
To ensure rigorous evaluation, unexecutable scripts are assigned a score of zero for DS, MJ and F1.

\subsection{Baselines}

\noindent \textbf{General MLLMs.}
We evaluate both closed-source and open-source MLLMs as the general-purpose baselines. The closed-source group includes GPT-4o \citep{gpt-4o-website}, GPT-4o-mini \citep{gpt-4o-mini-website}, GPT-5-mini \citep{gpt-5-website}, Claude-3.5-Sonnet \citep{claude-3-5-sonnet-website}, and Claude-Sonnet-4 \citep{claude-sonnet-4-website}. 
The open-source group covers 9 representative vision–language models including Qwen3-VL-2B, Qwen3-VL-4B, Qwen3-VL-8B \citep{qwen3technicalreport}, 
InternVL-3.5-2B, InternVL-3.5-4B, InternVL-3.5-8B \citep{wang2025internvl35advancingopensourcemultimodal}, 
DeepSeek-VL2-Tiny \citep{lu2024deepseekvl}, 
Phi-3.5-Vision-4B \citep{abdin2024phi3technicalreporthighly}, 
and LLaVA-v1.6-7B \citep{liu2023visual}.

\noindent \textbf{Chart MLLMs.}
We also compare against chart-specialized MLLMs.
ChartLlama-13B \citep{han2023chartllama} adapts LLaVA to chart reasoning via instruction tuning.
ChartMoE-8B \citep{xu2025chartmoe} advances chart understanding through a mixture-of-experts multimodal projector.
ChartCoder-7B \citep{zhao-etal-2025-chartcoder} directly targets chart-to-code generation by employing a code LLM as its language backbone.

\subsection{Main Results}

Existing MLLMs often exhibit pronounced disparities in chart-to-code generation across different programming languages, as illustrated in Table \ref{table:experimental_result}.
ChartCoder-7B achieves 96.2 ER and 86.4 MJ on the Python subset of Chart2NCode, while its performance deteriorates significantly on other languages, yielding 17.9 ER on LaTeX and zero valid generations for R.
We observe similar performance gaps in general-purpose open-source models on the Chart2NCode test set. 
DeepSeek-VL2-3B and Phi-3.5-Vision-4B display acute imbalances, achieving around 70 ER on Python while falling below 10 ER on LaTeX. 
Qwen3-VL-8B suffers from severe fidelity degradation, showing markedly reduced visual alignment for R (57.2 MJ) and LaTeX (57.1 MJ) compared to Python (80.8 MJ). 
Proprietary models display this same tendency in a more moderate form; GPT-4o-mini and Claude-Haiku-3.5 maintain the DS score above 80 on Python, but drop to around 60 DS on LaTeX.

CharLuMA effectively addresses cross-language disparities, establishing itself as the open-source MLLM for universal chart-to-code generation.
On established Python benchmarks, CharLuMA-6.7B delivers top-tier results with 92.3 ER and 77.4 DS on ChartMimic, along with 96.2 ER and 68.3 DS on Plot2Code.
CharLuMA-1.3B shows high efficiency, securing 83.0 ER and 71.8 DS on ChartMimic.
On the Chart2NCode test set, CharLuMA-6.7B sustains balanced performance across languages, achieving 88.7 DS and 83.5 F1 on Python, 81.8 DS and 80.9 MJ on R, and 72.5 DS and 74.2 MJ on LaTeX.
Notably, it rivals Claude-Haiku-3.5 and GPT-4o-mini on all benchmarks, significantly narrowing the gap with proprietary systems.

%% file: sections/6_analysis.tex
\section{Further Study}
\label{sec:ablation_study}

We conduct ablation studies and analyses to disentangle component contributions, demonstrating CharLuMA's robustness and interpretability.

\subsection{Model Architecture Ablation}
\label{sec:ablation_archi}

We conduct ablation studies on CharLuMA-1.3B with the Chart2NCode test set to examine alternative architectures, backbone choices, and subspace–router configurations.

\begin{table}[t]
    \belowrulesep=0pt
    \aboverulesep=0pt
    \renewcommand{\arraystretch}{1.2}
    \setlength{\belowcaptionskip}{0.1cm}
    \centering
    \setlength\tabcolsep{3pt}
    \scalebox{0.80}{
    \begin{tabular}{p{1.0cm}<{\centering}|p{3.2cm}<{\centering}|p{1.2cm}<{\centering}p{1.2cm}<{\centering}p{1.2cm}<{\centering}}
        \toprule
         \multirow{2}{*}{\makecell[c]{Model \\ Size}} & \multirow{2}{*}{\makecell{Projector \\ Architecture}}  & \multicolumn{3}{c}{\textbf{Chart2NCode}} \\
        \cmidrule(r){3-5}
         & & ER  & DS & MJ \\
        \cmidrule(r){1-5}
        \multirow{3}{*}{1.3B} & Linear MLP & \underline{88.1}    & \underline{76.9}  & \underline{69.5} \\
        & Mixture-of-MLP & 87.9      & 75.1    & 68.2 \\
        & Subspace Adapter & \textbf{91.1} & \textbf{78.9}  & \textbf{72.3}\\
        \midrule
        \multirow{3}{*}{6.7B} & Linear MLP & 91.0   & \underline{78.2}  & \underline{76.3}   \\
        & Mixture-of-MLP & \underline{91.9}   & 77.4     & 76.8 \\
        & Subspace Adapter & \textbf{94.5} & \textbf{81.0} & \textbf{81.1} \\
        \bottomrule
    \end{tabular}
    }
    \caption{Performance of alternative multimodal projector architectures during the instruction tuning stage of CharLuMA-1.3B and -6.7B on the Chart2NCode test set. Results are averaged over all three languages.} 
    \label{table:ablation_model_archi}
\end{table}

\begin{table}[t]
    \belowrulesep=0pt
    \aboverulesep=0pt
    \renewcommand{\arraystretch}{1.2}
    \setlength{\belowcaptionskip}{0.1cm}
    \centering
    \setlength\tabcolsep{3pt}
    \scalebox{0.75}{
    \begin{tabular}{p{4.0cm}<{\centering}|p{1.5cm}<{\centering}|p{1.2cm}<{\centering}p{1.2cm}<{\centering}p{1.2cm}<{\centering}}
        \toprule
         \multirow{2}{*}{\makecell[c]{Language Model}} & \multirow{2}{*}{\makecell{Vision \\ Encoder}}  & \multicolumn{3}{c}{\textbf{Chart2NCode}} \\
        \cmidrule(r){3-5}
         & & ER & DS & MJ  \\
        \cmidrule(r){1-5}
        DeepSeek-LLM-7B & SigLIP & 88.6 & 77.1 & 74.3  \\
        DeepSeek-Coder-6.7B & CLIP & \underline{88.8} & \underline{79.2} & \underline{75.0} \\
        DeepSeek-Coder-6.7B & SigLIP & \textbf{94.5} & \textbf{81.0} & \textbf{81.1} \\
        \bottomrule
    \end{tabular}
    }
    \caption{Ablation study of backbone choices in CharLuMA on the Chart2NCode test set, with results averaged over all three languages.} 
    \label{table:ablation_base_model}
\end{table}

\begin{table}[t]
    \belowrulesep=0pt
    \aboverulesep=0pt
    \renewcommand{\arraystretch}{1.1}
    \setlength{\belowcaptionskip}{0.1cm}
    \centering
    \setlength\tabcolsep{3pt}
    \scalebox{0.75}{
    \begin{tabular}{p{1.5cm}<{\centering}p{1.5cm}<{\centering}p{1.5cm}<{\centering}|p{1.2cm}<{\centering}p{1.2cm}<{\centering}p{1.2cm}<{\centering}}
        \toprule
        \multirow{2}{*}{\makecell[c]{Total \\ Subspace}} & \multirow{2}{*}{\makecell[c]{Activated \\ Subspace}} & \multirow{2}{*}{\makecell[c]{Total \\ Router}} & \multicolumn{3}{c}{\textbf{Chart2NCode}} \\
        \cmidrule(r){4-6} 
         & & & ER & DS & MJ \\
        \midrule
        16 & 8 & 3 & 88.9           & 77.6         & 70.5  \\
        32 & 8 & 3 & \underline{89.4}           & 77.8          & \underline{71.3}\\
        64 & 32 & 3 & 87.8           & 75.6         & 68.5  \\
        \cmidrule(l){1-3}
        32 & 16 & 1 & 86.1          & 75.1          & 67.0  \\
        32 & 32 & 0 & 85.8          & 73.2          &  66.3  \\
        \cmidrule(l){1-3}
        32 & 16 & 3 & \textbf{91.1} & \textbf{78.9} & \textbf{72.3} \\
        \multicolumn{3}{c|}{\textit{w/o warming up before finetuning}} & 87.1          & 75.6          & 67.9 \\
        \multicolumn{3}{c|}{\textit{w/o freezing A matrix of adapter}} & 90.2          & \underline{78.0}          & 70.1    \\
        \bottomrule
    \end{tabular}
    }
    \caption{Ablation study of subspace-router configurations in CharLuMA-1.3B on the Chart2NCode test set, with results averaged over all three languages.}
    \label{table:ablation_router_subspace}
\end{table}

\noindent \textbf{Alternative Architecture.} 
We compare our low-rank \textit{subspace adapter} with two alternative projector designs in Table \ref{table:ablation_model_archi}. 
The \textit{linear MLP} baseline \citep{belouadi2024detikzify, belouadi2024automatikz, zhao-etal-2025-chartcoder} yields modest improvements, resulting in 88.1 ER and 69.5 MJ for the 1.3B model.
The \textit{Mixture-of-MLP} design \citep{10.5555/3737916.3742086, xu2025chartmoe} replaces the MLP block with a sparsely gated mixture-of-experts, each initialized from a pretrained MLP block, and we adapt it with a hard-routing policy that activates the language-specific and shared experts (see Appendix \ref{sec:analysis_setting}). This achieves 91.9 ER and 76.8 MJ for the 6.7B model.
In contrast, our low-rank \textit{subspace adapter} achieves the strongest results across both model sizes with a far more compact design compared to Mixture-of-MLP.

\noindent \textbf{Backbone Choices.}
We examine the effect of language and vision backbones in Table~\ref{table:ablation_base_model}. 
Our default configuration with DeepSeek-Coder-6.7B and SigLIP yields the strongest results.
Deviating from this setup leads to consistent performance drops: replacing the language model with DeepSeek-LLM-7B lowers scores to 88.6 ER and 74.3 MJ, while substituting the vision encoder with CLIP-Large reduces performance to 88.8 ER and 75.0 MJ.

\noindent \textbf{Subspace–Router Configurations.} 
We compare different subspace settings, routing strategies and training choices in CharLuMA-1.3B on the Chart2NCode test set in Table \ref{table:ablation_router_subspace}.
Regarding subspace settings, we identify our default 32–16 configuration (Row 6) as the optimal balance; it outperforms both smaller and larger total subspaces (Rows 1 and 3) and surpasses the insufficient active capacity of the 32–8 setting (Row 2).
We further confirm the necessity of language-guided routing in Rows 4–6,
where replacing language-specific routers with a single shared router reduces DS from 78.9 to 75.1, and removing them entirely lowers it to 73.2. 
Finally, Rows 7–8 validate our training choices: removing the warming-up stage destabilizes specialization and lowers DS to 75.6, while unfreezing the $\mathbf{A}$ matrix compromises the low-rank representation, reducing DS to 78.0.

\subsection{Language Structure Ablation}
\label{sec:ablation_lang}

We examine the impact of language structure during training on Chart2NCode. To guarantee fairness, we enforce a fixed budget of training samples across all settings.
Under the single-language setting, we give script supervision for each image from one single language.
For the two-language setting, we randomly sample half of the charts and provide script supervision in two different languages (\textit{e.g.}, pairing each image with both LaTeX and Python scripts).
Similarly, in three-language setting, we randomly sample one-third of the charts and apply script supervision using all three languages.
Additionally, we include an unaligned baseline using only the original source script per chart, which biases towards predominant languages and lacks multi-view alignment. Finally, the number of routers is configured to match the target language count (see Appendix~\ref{sec:analysis_setting}).

Figure~\ref{figure:ablation_router_language} shows that greater language diversity leads to substantial improvements on the Chart2NCode test set, effectively outweighing the reduction in unique visual exposure. The three-language model achieves the highest execution rates and visual fidelity scores across all languages, significantly outperforming single- and two-language models.
Notably, training on unaligned source data induces systematic biases that impede universality, skewing the model toward the dominant language, \textit{i.e.}, Python. This positions Chart2NCode as the first dataset to provide the diverse, aligned supervision necessary for robust chart-to-code generation.

\begin{figure}
    \centering
    \includegraphics[width=\columnwidth]{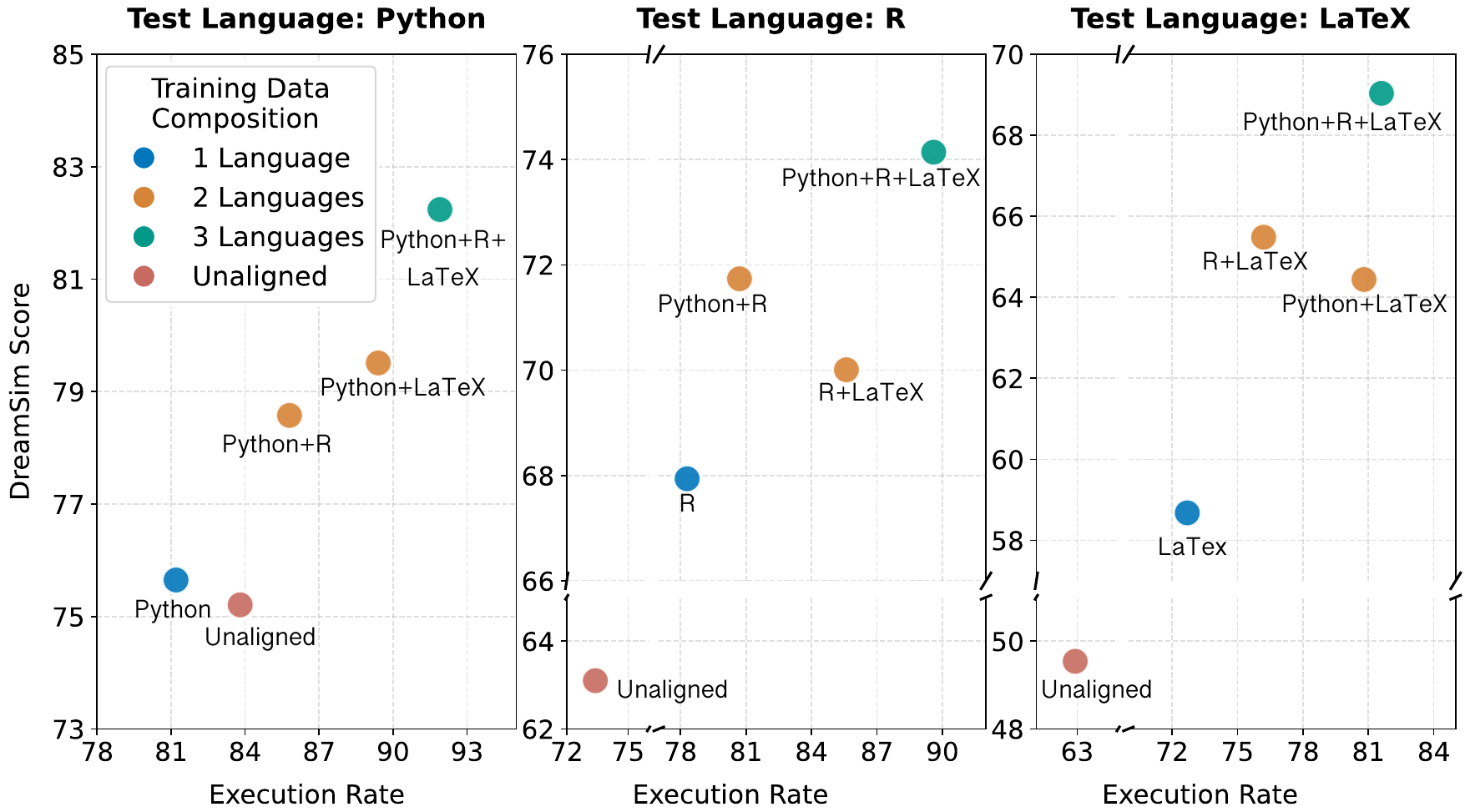}
    \captionof{figure}{Ablation study of language structure using CharLuMA-1.3B on the Chart2NCode test set.}
    \label{figure:ablation_router_language}
\end{figure}

\subsection{Subspace Activation Analysis}

\begin{figure}
    \centering
    \includegraphics[width=\columnwidth]{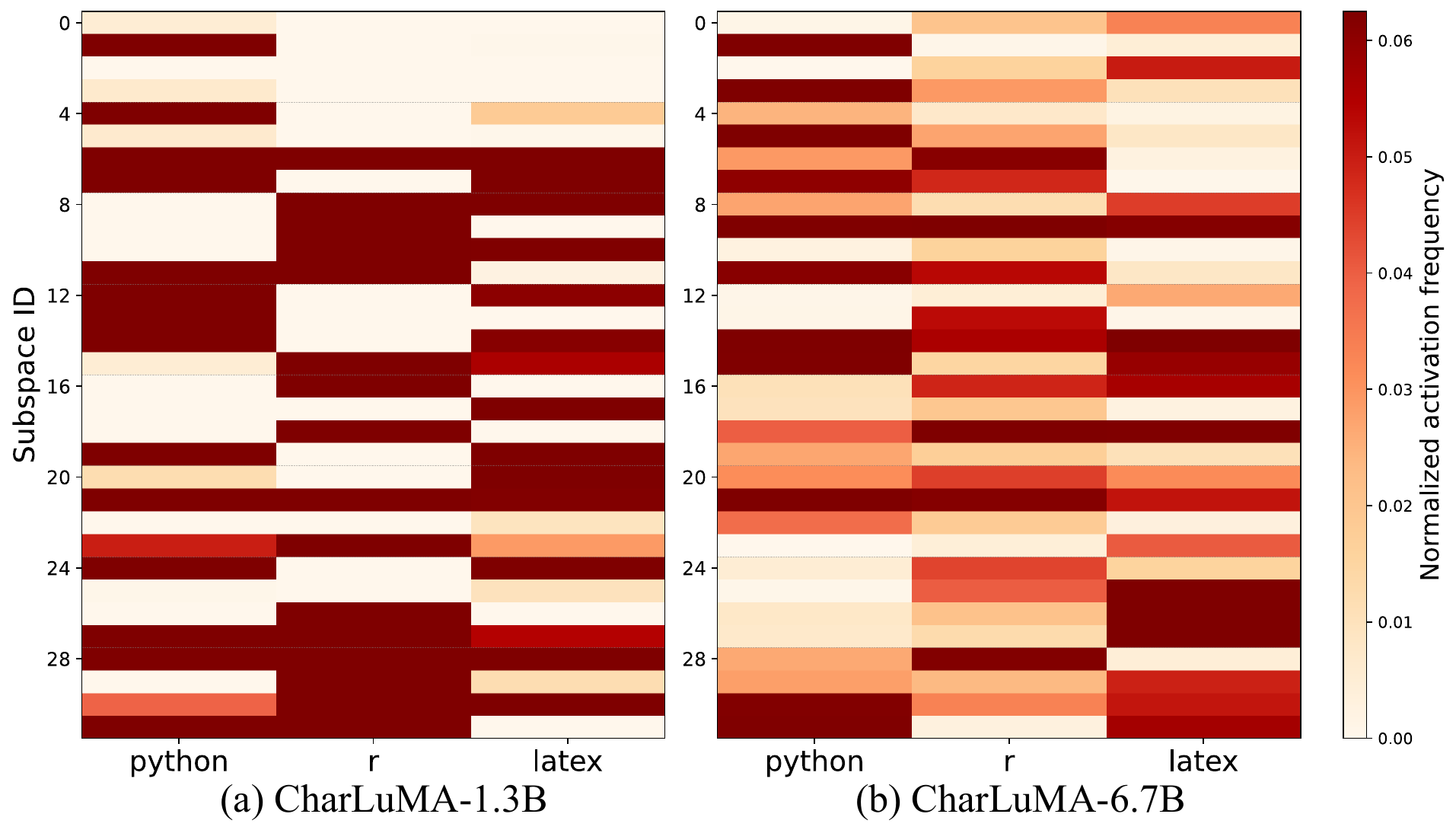}
    \caption{Heatmap of subspace activation frequency for (a) CharLuMA-1.3B and (b) CharLuMA-6.7B.}
    \label{figure:heatmap_subspace}
\end{figure}

We visualize the normalized activation frequency of 32 subspaces across CharLuMA's language-specific routers in Figure~\ref{figure:heatmap_subspace}. These heatmaps reveal a hybrid allocation strategy, featuring compact shared clusters alongside broader language-specific zones. 
In CharLuMA-1.3B, subspaces 21, 23, and 30 are frequently activated across all languages, while subspace 1 is used primarily for Python, 18 for R, and 17 for LaTeX. 
CharLuMA-6.7B shows a more balanced distribution, with most subspaces—such as 8, 20, and 29—exhibiting intermediate activation frequencies across the three languages. 
These findings confirm that the architecture facilitates smooth multi-language integration across various model scales.

\begin{figure}
    \centering
    \includegraphics[width=\columnwidth]{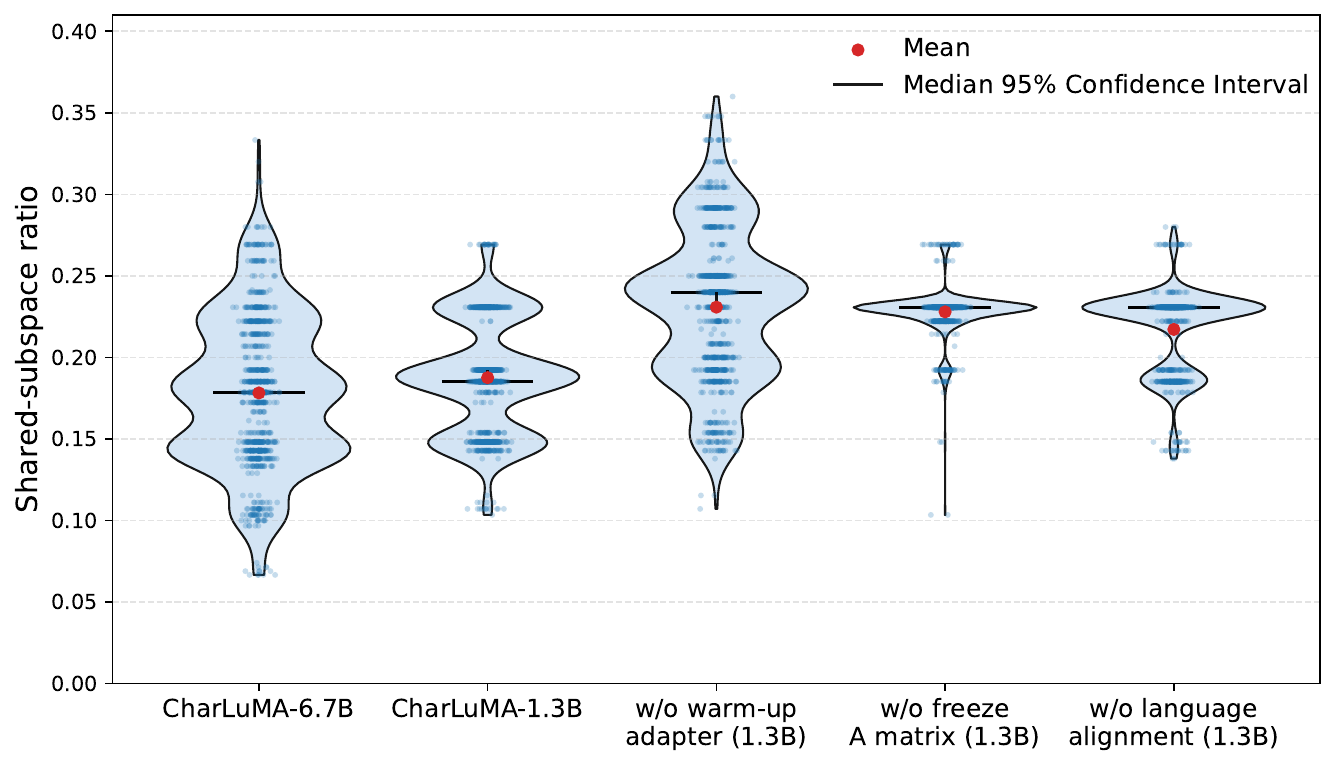}
    \captionof{figure}{Distribution of shared-subspace ratios across CharLuMA and ablated models.}
    \label{figure:shared_expert_ratio}
\end{figure}

We introduce the \textit{shared-subspace ratio} to quantify the cross-language allocation of subspaces, defined as the proportion of experts activated by all routers relative to the total activated set (Appendix \ref{sec:analysis_setting}).
Figure~\ref{figure:shared_expert_ratio} reports the distribution of this ratio over a random sample of 1,000 Chart2NCode instances. 
CharLuMA-1.3B achieves a median ratio of 0.19, corresponding to roughly 5 shared experts out of 27. CharLuMA-6.7B shows a similar pattern with a median of 0.18, where about 4.9 experts are shared out of 27.5 on average. 
This indicates that scaling preserves a compact shared core, while allocating the increased capacity to expanded language-specific subspaces.
In contrast, the ablated 1.3B variants exhibit inflated ratios (0.23–0.24), resulting from a contraction of the activation pool that compromises language-specific specialization.

\subsection{Quantitative Analysis}

We conduct a detailed error analysis of CharLuMA-6.7B on the Chart2NCode test set to reveal distinct language-specific failure dynamics. 
Execution errors in Python and R stem primarily from logic and data discrepancies, led by dimension mismatches (72.3\% in Python, 56.1\% in R) and undefined variables (11.9\% in Python, 22.0\% in R). 
LaTeX is uniquely prone to collapsing into unexecutable states due to rigid syntactic constraints (55.5\%) such as missing braces.
Beyond execution, successful generations often exhibit reproduction limitations relating to missing annotations, inaccurate subtypes, or stylistic inconsistencies (Appendix~\ref{sec:errror_analysis}).
Case studies in Appendix~\ref{sec:examples} further confirm CharLuMA's superior cross-language stability against GPT-4o and ChartCoder. 

%% file: sections/appendix.tex
\section{Dataset}

\subsection{Data Acquisition}
\label{sec:data_acquisition}

We collect single-language plotting scripts from established datasets and publicly available repositories governed by permissive licenses as our source data. ChartCoder \citep{zhao-etal-2025-chartcoder} contributes approximately 160k chart-to-Python scripts, while DaTikZ \citep{belouadi2024automatikz} provides 49k vector-graphics-to-LaTex scripts, of which 8.8k correspond to charts with explicit axis structures. In addition, we curated 40k R plotting scripts from Stack Overflow \footnote{Retrieved using StackAPI with keywords representative of R plotting functions and libraries, including \texttt{ggplot}, \texttt{plot\_ly}, \texttt{geom}, \texttt{plot(}, \texttt{hist}, \texttt{boxplot} and so on.}, strictly adhering to the platform's attribution requirements and CC BY-SA data usage policy. 
To handle deprecated or non-executable scripts, we employed GPT-4o as an automated debugging assistant, guided by the prompt instructions in Figure~\ref{fig:prompt_debug_execut}, with a total API cost of 132.2~USD.

\subsection{Annotation Pipeline}
\label{sec:annot_pipeline}

\noindent \textbf{Metadata Structure and Extraction.} 
We adopt a hierarchical metadata schema to capture chart information at three levels: figure, axis, and object. This structure provides a standardized representation of chart elements across languages while preserving both global properties and fine-grained graphical details. At the figure level, metadata records global properties such as the overall title, background color and legend, plot size (width, height, and units), twin-axis relationships, and subplot layout. For each axis, metadata focuses on type-agnostic attributes including axis titles, x- and y-axis labels, tick values and labels, legends, grids, panel boxes, background color, and annotations. At the object level, metadata captures fine-grained properties of graphical elements grouped into patches, lines, collections, and images. For each object, visual properties such as color, transparency, line width, marker style, and hatch patterns are recorded, together with precise geometric information such as rectangle bounds, circle centers and radii, polygon vertices, line coordinates, scatter offsets, and heatmap arrays. Cleaned labels are associated with color or stylish values where available, ensuring consistency with legends and categorical encodings.

Metadata is extracted by executing or parsing plotting scripts in their native environments. For Python plotting scripts, each script is executed in an isolated runtime, and the figure is inspected using \texttt{fig.get\_axes()}. Axis-level attributes are gathered through standard APIs such as \texttt{ax.get\_title()}, \texttt{ax.get\_xlabel()}, and \texttt{ax.get\_yticks()}. Object-level elements are obtained by iterating over \texttt{ax.patches}, \texttt{ax.lines}, \texttt{ax.collections} and so on. For R scripts based on \texttt{ggplot}, code is evaluated to collect the plotting object \texttt{p} built via \texttt{ggplot\_build()}. We extract axis-level metadata from structures such as \texttt{p\$labels\$title}, \texttt{p\$mapping\$y}, and \texttt{p\$theme\$panel.border}, while object-level metadata is obtained by iterating over \texttt{p\$layers}. For base R graphics, we wrap high-level functions like \texttt{barplot}, \texttt{hist}, and \texttt{boxplot}, as well as low-level commands such as \texttt{text}, \texttt{legend}, and \texttt{grid}, to capture metadata during execution. For LaTeX, we use the regular-expression parsing to detect \texttt{axis} environments while drawing commands are parsed to recover object geometries such as rectangles, circles, and paths.

\noindent \textbf{Template Design.} 
The templates are parameterized chart skeletons that translate extracted metadata into executable plotting code. Each template specifies placeholders for chart elements such as titles, axis labels, ticks, grids, legends, annotations, and objects, which are directly filled from metadata. The overall structure is consistent across languages, but implementation details differ. Taking the bar type for example, Python uses functions like \texttt{ax.bar} or \texttt{ax.barh} in matplotlib, R employs \texttt{geom\_bar} in ggplot, and LaTeX relies on declarative PGFPlots options such as \texttt{xbar}, \texttt{ybar} and \texttt{addplot} using TikZ. 

To maintain cross-language consistency during template instantiation, we employ an attribute-mapping process that normalizes visual properties across Python, R, and LaTeX. Legend locations are aligned so that values such as ``upper right'' in Python correspond to ``right'' in R and ``north east'' in LaTeX. Font styles are unified by mapping bold and italic settings into Python’s weight and style fields, R’s font face descriptors, or LaTeX commands like \texttt{bfseries} and \texttt{itshape}. Font sizes are standardized by converting numeric values in Python and R into LaTeX size categories such as \texttt{small} or \texttt{Large}. Annotation alignment is harmonized by translating Python’s top, bottom, and center into equivalent justification values in R and LaTeX. Marker and line styles are also consolidated through shared dictionaries, ensuring that a logical style such as circle, dashed, or cross is rendered consistently across all languages. This mapping guarantees that semantic attributes are preserved even when the syntax differs, allowing metadata extracted in one language to be instantiated in another without loss of fidelity.

\noindent \textbf{Metadata-Template Matching.}
A critical step in our automatic pipeline is to identify the correct template once the metadata of a chart has been extracted. We address this by assigning each chart a type and subtype based on patterns in the object-level metadata. Taking bar charts for example, we examine the geometry of rectangular patches: overlapping intervals reveal stacked bars, repeated clusters of equal size indicate grouped bars, with other cases default to base bars. For pie charts, subtype inference is based on patch geometry and offsets: the presence of an inner radius or nonzero x position signals a donut chart, displaced segment centers indicate exploded pies, and their combination yields donut–exploded pies. These inference rules allow the system to automatically select the most appropriate template across diverse chart variants.

\noindent \textbf{LLM-assisted Debugging.} 
We incorporate an LLM-assisted debugging module based on GPT-4o to handle cases where no suitable template is identified or when an instantiated template fails to execute. Instruction prompts for these two scenarios are provided in Figure~\ref{fig:prompt_debug_missing} and Figure~\ref{fig:prompt_debug_execut}. The total expenditure on the OpenAI API is 316.6~USD.

\subsection{Quality Assurance}
\label{sec:data_quality_check}

We conduct a human evaluation to systematically assess the cross-language fidelity of Chart2NCode. We randomly sample 1,000 chart–Python–R–LaTeX quadruples from the Chart2Ncode dataset, which are independently annotated by three annotators. All annotators were recruited on campus, with eligibility requiring prior experience in data visualization and programming in Python, R, and LaTeX. They were compensated in accordance with the institution’s standard remuneration policies for academic work.

\begin{figure*}[t]
    \centering
    \includegraphics[width=0.95\linewidth]{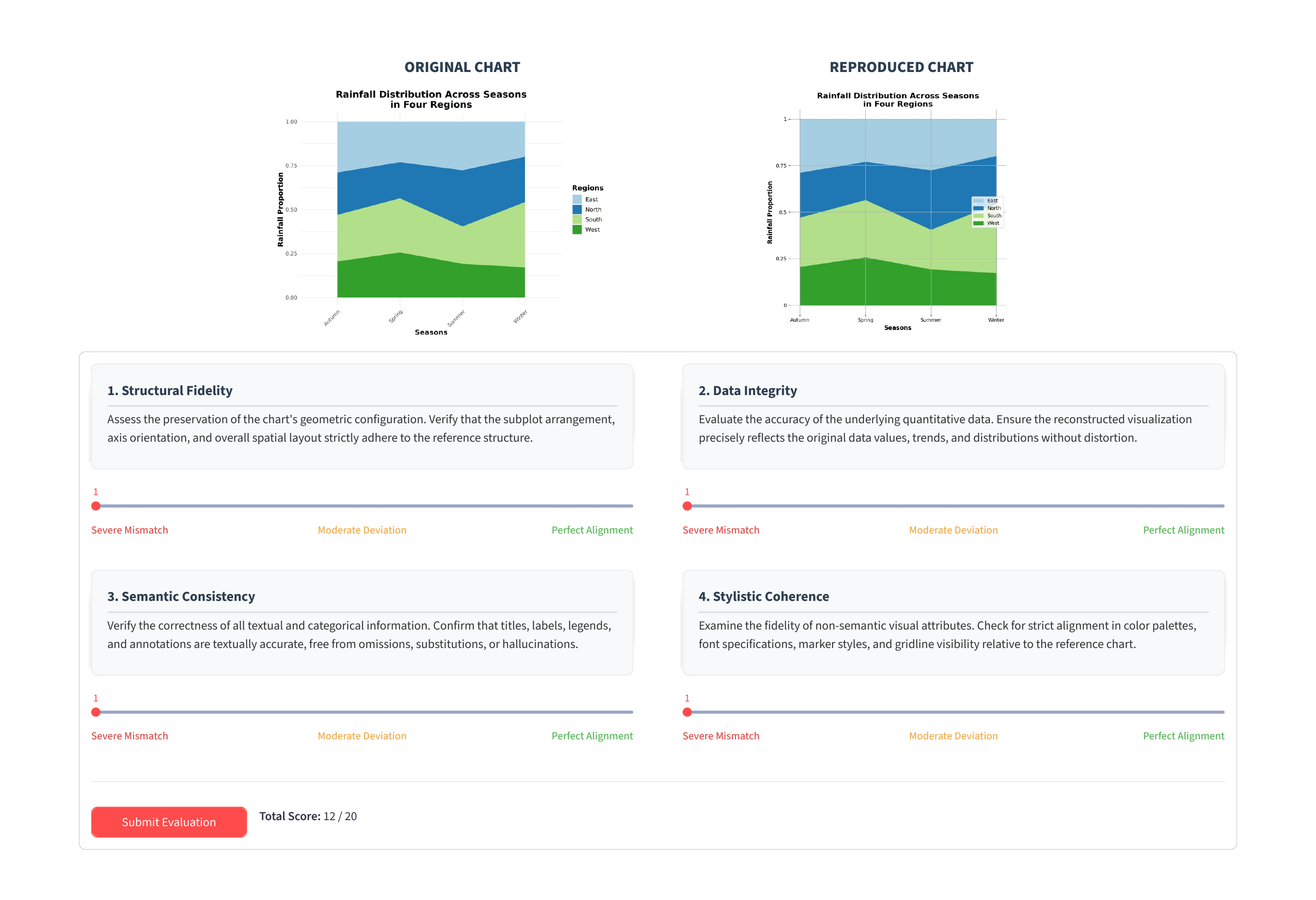}
    \caption{Screenshot of the human quality checking questionnaire.}
    \label{fig:questionnaire_human_quality}
\end{figure*}

We conduct evaluations for each quadruple, comparing the reproduced charts in Python, R, and LaTeX against the original image, and annotators assess their fidelity across four dimensions. \textit{Structural fidelity} measures whether the geometric arrangement of the chart is preserved, including the number and configuration of subplots and axis orientation. \textit{Data integrity} evaluates whether the underlying quantitative values are reproduced exactly, meaning that the reconstructed chart reflects the same data table as the original. \textit{Semantic consistency} assesses whether textual and categorical information is maintained, ensuring that titles, axis labels, legends, and annotations convey the same meaning without omissions, substitutions, or hallucinations. \textit{Stylistic coherence} concerns the visual presentation, requiring that non-semantic design elements—such as color palettes, font attributes, line styles, and grid line visibility—remain consistent with the original chart. All dimensions are rated on a 1–5 scale, where 1 denotes severe mismatch and 5 denotes perfect alignment.
A screenshot of the evaluation interface is available in Figure \ref{fig:questionnaire_human_quality}.

\begin{table}
\centering
\small
\scalebox{0.95}{
\begin{tabular}{lcccc}
\toprule
\textbf{Dimension} & \textbf{Ann.~1} & \textbf{Ann.~2} & \textbf{Ann.~3} & \textbf{Avg.} \\
\midrule
Structural fidelity   & 98.7 & 97.8 & 98.1 & 98.2 \\
Data integrity        & 94.5 & 95.8 & 95.3 & 95.2 \\
Semantic consistency  & 97.9 & 96.6 & 97.7 & 97.4 \\
Stylistic coherence   & 96.2 & 95.7 & 95.5 & 95.8 \\
\bottomrule
\end{tabular}
}
\caption{Proportion (\%) of examples with average rating $\ge 4$ on 1,000 sampled quadruples, reported per annotator and averaged across annotators. Overall row averages the four dimensions.}
\label{tab:human_eval_prop}
\end{table}

We compute the average per-dimension score across annotators for each example, and report the proportion of examples achieving an average score of at least 4. As shown in Table~\ref{tab:human_eval_prop}, the evaluation results confirm high fidelity across dimensions: 
98.2\% of examples exceed the threshold for structural fidelity, 95.2\% for data integrity, 97.4\% for semantic consistency, and 95.8\% for stylistic coherence. 
To rigorously assess reliability on the full ordinal scale, we compute Krippendorff's $\alpha$ \citep{krippendorff2011computing}. The resulting average $\alpha$ of 0.81 indicates substantial agreement beyond chance, representing a strong and practical level of consistency for human judgment in chart reproduction tasks.

\subsection{Detailed Data Staistics}
\label{sec:detailed_data_stat}

We report detailed statistics for the Chart2NCode dataset, spanning chart type distributions and code complexity. To ensure a robust training and evaluation source,  the Chart2NCode dataset covers 20 distinct chart categories. Table \ref{tab:test_set_dist} details the frequency and percentage of each type. The dataset explicitly challenges models with advanced composite visualizations, including Multidiff, which requires generating multiple heterogeneous subplots, and Combination, which involves overlaying distinct geometric types on a shared coordinate system.

Regarding code complexity on the Chart2NCode dataset, we utilized the Llama 3 tokenizer~\cite{llama-3-website} to calculate token counts. The resulting statistics show a mean length of 384.1 tokens for Python ($\sigma=189.7$, median 348.0), 591.8 tokens for R ($\sigma=242.0$, median 545.0), and 637.1 tokens for LaTeX ($\sigma=247.1$, median 595.0).

\begin{table}[ht]
    \centering
    \renewcommand{\arraystretch}{1.3}
    \scalebox{0.8}{
    \begin{tabular}{l|cccc}
        \hline
        Type & Area & Bar & Box & Bubble \\
        Percent & 5.5\% & 11.6\% & 5.3\% & 2.0\% \\
        \hline
        Type & Density & Donut & ErrorBar & ErrorPoint \\
        Percent & 1.9\% & 3.2\% & 2.9\% & 4.8\% \\
        \hline
        Type & Heatmap & Histogram & Line & Lollipop \\
        Percent & 6.9\% & 1.3\% & 12.4\% & 0.5\% \\
        \hline
        Type & Pie & Quiver & Radar & Scatter \\
        Percent & 7.6\% & 0.7\% & 6.7\% & 6.5\% \\
        \hline
        Type & Violin & 3D & Multidiff & Combination \\
        Percent & 6.3\% & 1.0\% & 9.1\% & 3.8\% \\
        \hline
    \end{tabular}
    }
    \caption{Distribution of chart types in Chart2NCode.}
    \label{tab:test_set_dist}
\end{table}

\subsection{Case Study of Annotation Pipeline}
\label{sec:case_study}

\begin{figure*}[t]
    \centering
    \includegraphics[width=0.95\linewidth]{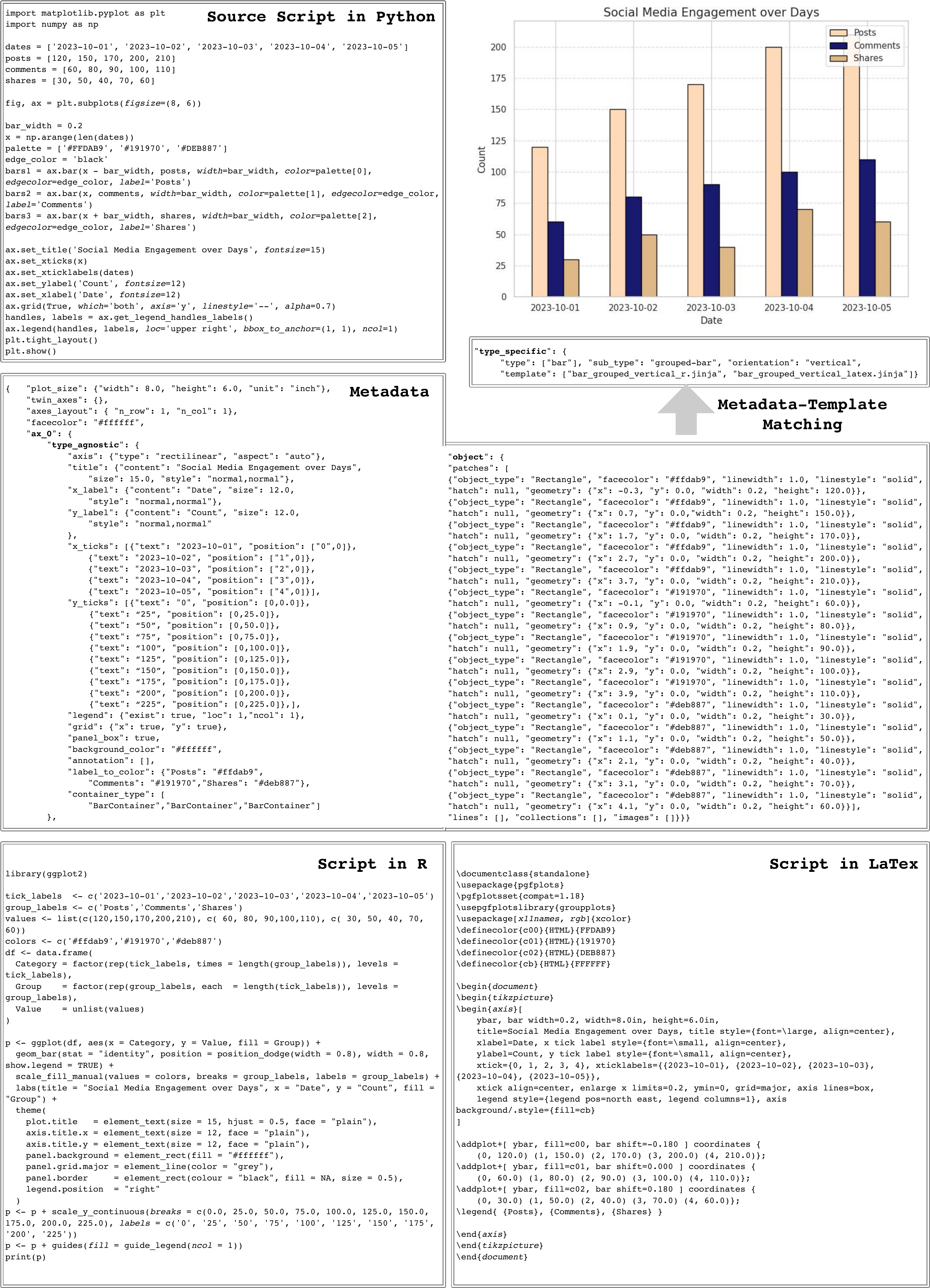}
    \caption{Case study of annotation pipeline in a vertical grouped bar chart.}
    \label{fig:annot_case_1}
\end{figure*}

\begin{figure*}[t]
    \centering
    \includegraphics[width=0.95\linewidth]{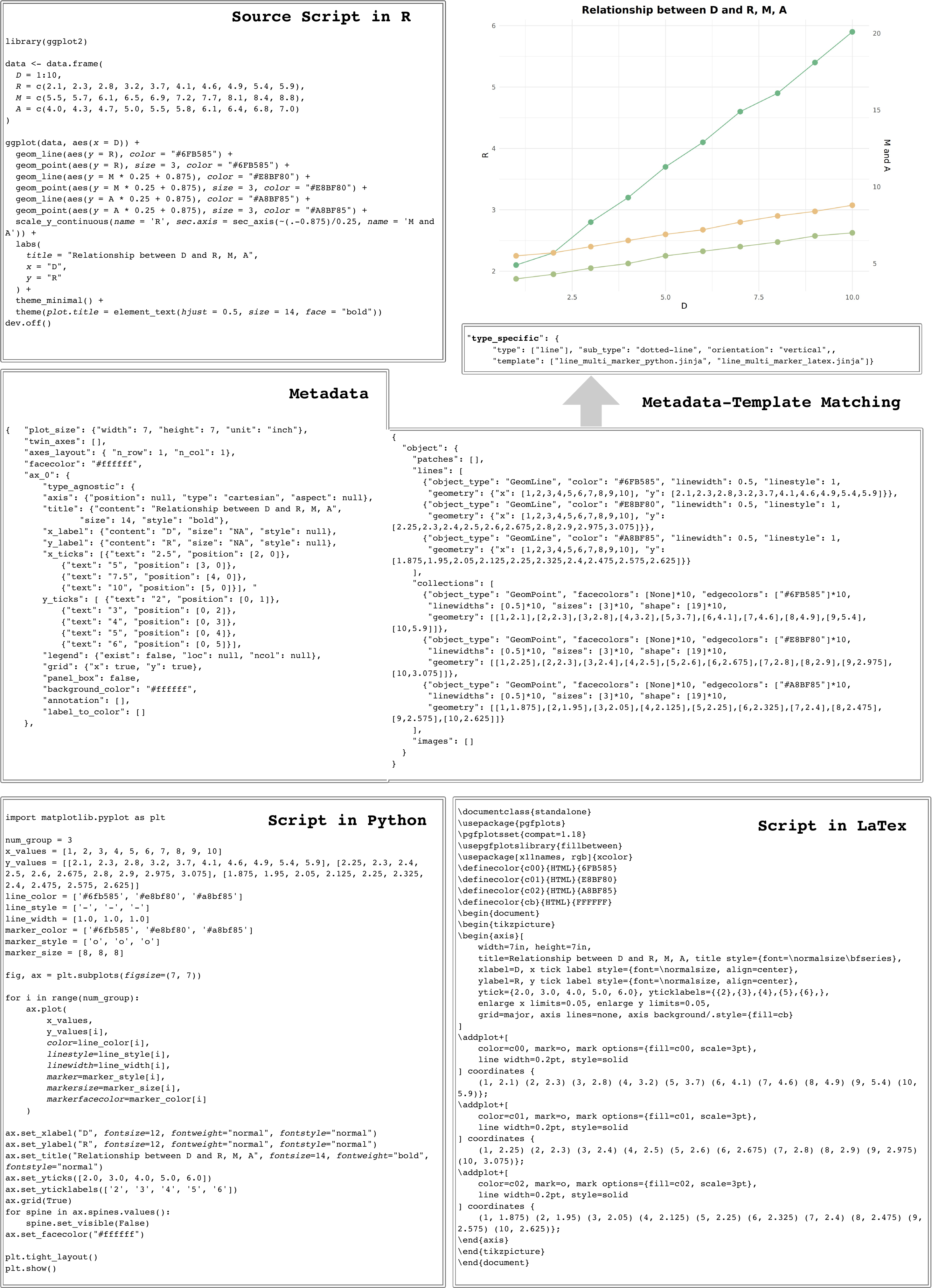}
    \caption{Case study of annotation pipeline in a dotted line chart.}
    \label{fig:annot_case_2}
\end{figure*}

We present two illustrative cases in Figure~\ref{fig:annot_case_1} and Figure~\ref{fig:annot_case_2} to demonstrate the functionality of our annotation pipeline.

\section{Experimental Settings and Results}

\subsection{Training and Evaluation Settings}
\label{sec:hyperparam_setting}

We adopt SigLIP \citep{zhai2023sigmoid} as the vision encoder and DeepSeek-Coder \citep{guo2024deepseekcoderlargelanguagemodel} as the LLM backbone, yielding two variants of our model: CharLuMA-1.3B and CharLuMA-6.7B. The multimodal projector is implemented as a standard two-layer MLP block augmented with our low-rank subspace adapter.

For alignment pretraining, we train the MLP block for one epoch on 900k chart–JSON pairs from ChartMoE-Align \citep{xu2025chartmoe}, while freezing both the vision encoder and LLM, with a learning rate of 2e-4. During instruction tuning, we first warm up the subspace pool and language-specific routers for 274 steps, and then perform full fine-tuning of the LLM backbone for one epoch on 175k chart–Python–R–LaTeX quadruples from Chart2NCode. In this stage, the vision encoder and MLP block remain frozen, the adapter is updated, and the learning rates are set to 2e-4 for warm-up and 2e-5 for fine-tuning. The low-rank projector  $\mathbf{A}$ remains frozen throughout. Each training batch is constructed to include all three languages.

All training experiments are conducted with a global batch size of 128 on 8× NVIDIA L40S GPUs. The training cost for CharLuMA-1.3B is approximately 82 GPU hours, consisting of 35 GPU hours for pretraining, 6 GPU hours for warm-up, and 41 GPU hours for fine-tuning. For CharLuMA-6.7B, the total cost is about 321 GPU hours, including 109 GPU hours for pretraining, 18 GPU hours for warm-up, and 193 GPU hours for fine-tuning. More training hyperparameters are in Table~\ref{tab:hyperparam_setting}.

For evaluation, we follow a standardized setup across all baselines, fixing the maximum token length to 2,048. The prompting format for the chart-to-code generation task is shown in Figure~\ref{fig:prompt_chart2code}, adapted from \citet{shi2024chartmimicevaluatinglmmscrossmodal}. Proprietary MLLMs evaluated include \texttt{gpt-4o-2024-08-06}, \texttt{gpt-4o-mini-2024-07-18}, \texttt{gpt-5-mini-2025-08-07}, \texttt{claude-3-5-haiku-20241022}, and \texttt{claude-sonnet-4-20250514}, all accessed through their official APIs. For open-source MLLMs, we directly run released checkpoints on NVIDIA L20 GPUs. Additionally, the total expenditure for MLLM-as-Judge metrics through the OpenAI API is 217.6~USD. 

\begin{table}[ht]
\centering
\small
\scalebox{0.85}{
\begin{tabular}{lccc}
\toprule
Hyperparameter & \makecell[c]{Alignment \\ Pretraining} & Warm-up & \makecell[c]{Instruction \\ Tuning} \\
\midrule
Learning rate & 2e-4 & 2e-4 & 2e-5 \\
LR schedule & Cosine decay & Cosine decay & Cosine decay \\
Optimizer & AdamW & AdamW & AdamW \\
Max tokens & 2,048 & 2,048 & 2,048 \\
Vision encoder & Frozen & Frozen & Frozen \\
LLM & Frozen & Frozen & Trainable \\
MLP Block & Trainable & Frozen & Frozen \\
Adapter & Frozen & Trainable & Trainable \\
\bottomrule
\end{tabular}
}
\caption{Training hyperparameters for CharLuMA across stages in Section \ref{sec:implementation_details}.}
\label{tab:hyperparam_setting}
\end{table}

\subsection{Detailed MLLM-as-Judge Metric}
\label{sec:mllm_judge}

We employ an MLLM-as-Judge (MJ) approach to assess visual alignment of reproduced charts. 
Following \citet{shi2024chartmimicevaluatinglmmscrossmodal}, we utilize GPT-4o \citep{gpt-4o-website} to quantify the extent to which a generated chart corresponds to the ground truth. Specifically, the generated chart and the ground-truth chart are both input into GPT-4o. The model is instructed to evaluate their similarity across six dimensions—text, layout, chart type, data integrity, style, and clarity—according to the criteria detailed in Figure~\ref{fig:prompt_gpt_scoring}. Subsequently, GPT-4o assigns a final similarity score ranging from 0 to 100.

\begin{figure*}[t]
    \centering
    \includegraphics[width=0.95\linewidth]{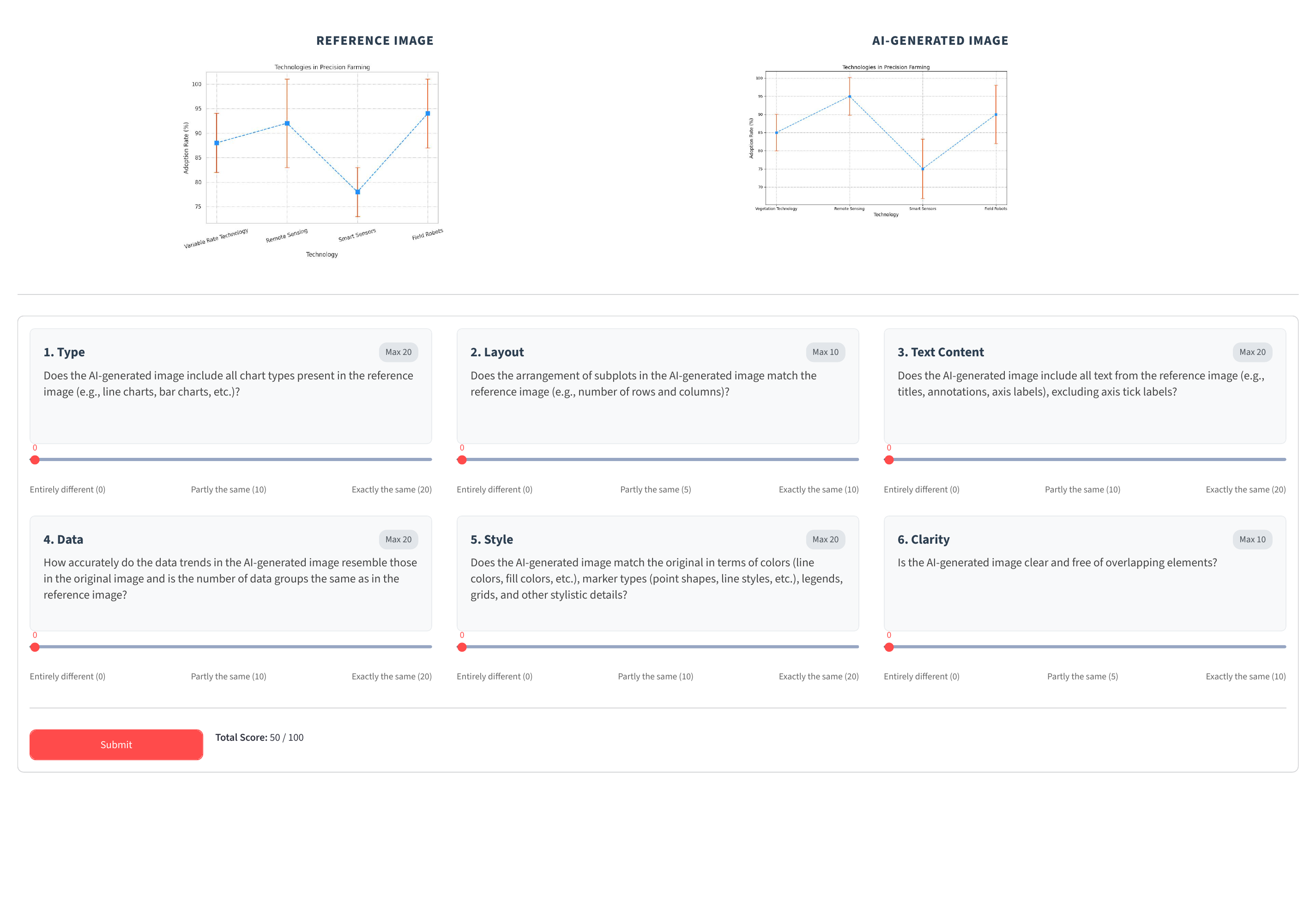}
    \caption{Screenshot of the human evaluation questionnaire for MLLM-as-judge metrics.}
    \label{fig:questionnaire_mllm_judge}
\end{figure*}

To validate the reliability of the MLLM-as-Judge metric, we analyze its correlation with human judgment. We select a subset of 100 examples from the Chart2NCode test set and gather the outputs of ChartLuMA-6.7B in Python, R, and LaTeX, resulting in a total of 300 figures for assessment. Three independent annotators were recruited based on their expertise in data visualization and proficiency in the relevant programming languages. They conducted the human evaluation using the interface in Figure~\ref{fig:questionnaire_mllm_judge}, which strictly mirrors the criteria in Figure~\ref{fig:prompt_gpt_scoring}. The final human score for each chart is derived by averaging the ratings from the three annotators. We calculate the Pearson correlation coefficient \citep{shi2024chartmimicevaluatinglmmscrossmodal} between the MLLM-as-Judge scores and the human evaluations, yielding a value of 0.7387. This strong correlation demonstrates that the MLLM-as-Judge approach serves as a reliable proxy for human visual assessment.

To verify the stability of the MLLM-as-Judge metric, we conduct five independent evaluation runs using the outputs of ChartLuMA-6.7B on the Chart2NCode test set. This yields consistent mean scores for Python (88.2), R (80.9), and LaTeX (74.3), with negligible standard deviations of 0.07, 0.06, and 0.08, respectively. These results confirm the high stability of MLLM-as-Judge metric.

To further address concerns regarding the reproducibility of closed-source APIs, we compare our primary judge (GPT-4o) against two leading open-source alternatives: Qwen3-VL-8B \citep{qwen3technicalreport} and InternVL3.5-8B \citep{wang2025internvl35advancingopensourcemultimodal}. Using the same subset of 100 examples and identical scoring criteria (Figure~\ref{fig:prompt_gpt_scoring}), we observe high Pearson correlations between the open-source judges and GPT-4o (0.8728 for Qwen3-VL-8B and 0.8540 for InternVL3.5-8B). Furthermore, both models demonstrate high alignment with human annotators (achieving correlations of 0.6975 and 0.6733 respectively). These results confirm that our evaluation protocol is robust and can be reliably reproduced using accessible open-source weights.

\subsection{Detailed Analysis Setting}
\label{sec:analysis_setting}

\begin{figure}
    \centering
    \includegraphics[width=\columnwidth]{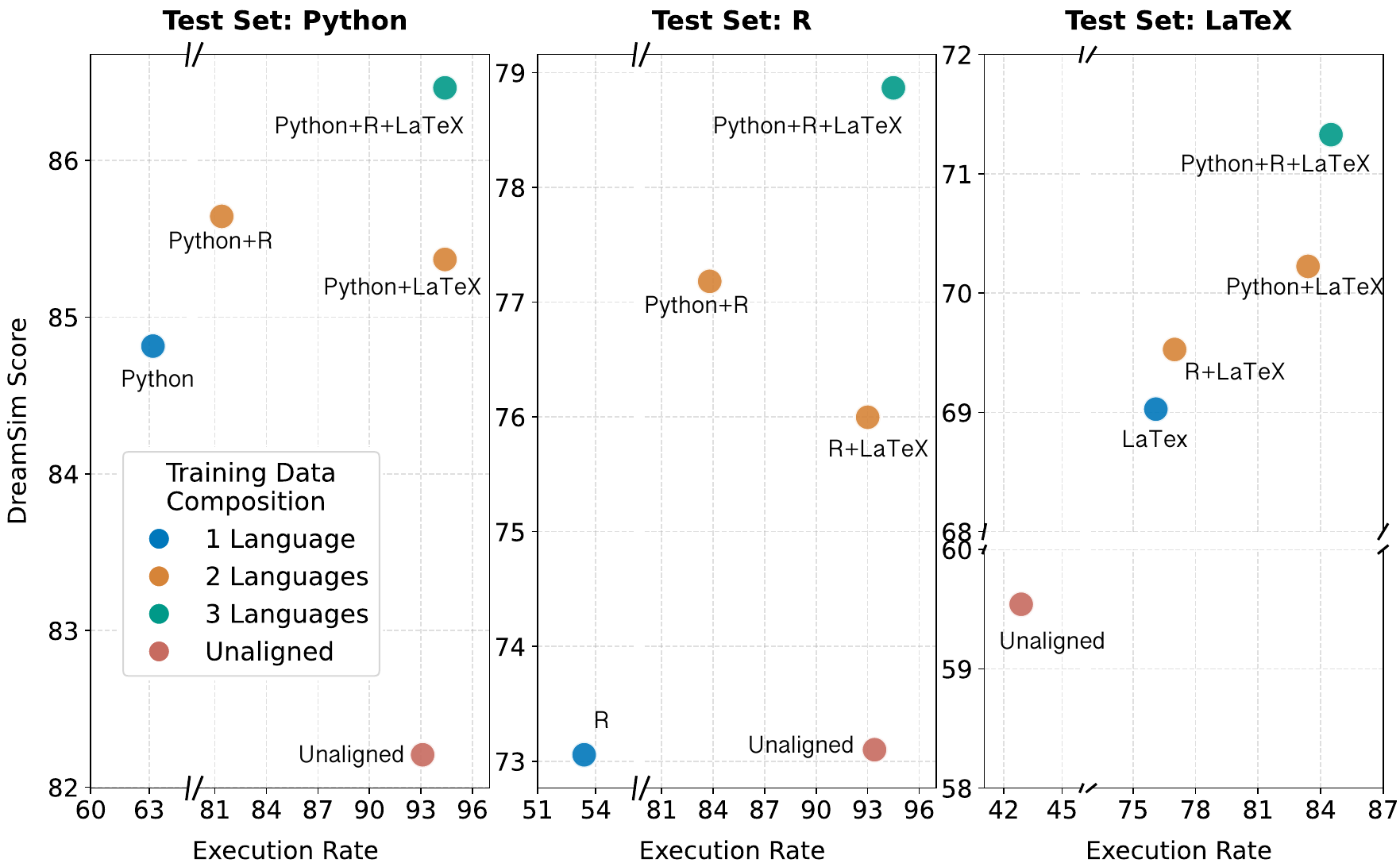}
    \captionof{figure}{Ablation study of language structure using CharLuMA-1.3B on the Chart2NCode test set.}
    \label{figure:ablation_router_language_appendix}
\end{figure}

\noindent \textbf{Alternative Architecture.} 
We compare our language-guided low-rank subspace adapter with two alternative connector architectures: a linear MLP and a Mixture-of-MLP. In the linear MLP setting, the pretrained MLP block, initialized on chart–JSON pairs, is directly fine-tuned on Chart2NCode. In the Mixture-of-MLP setting, four experts are initialized from the pretrained MLP block, one of which is frozen as a shared expert, while the remaining three serve as language-specific experts. Hard routing is applied such that, in a Python generation task, the Python-specific expert is activated jointly with the shared expert. This setup mirrors the configuration with four experts in total, of which two are activated for each time, as reported in prior studies \cite{10.5555/3737916.3742086, xu2025chartmoe}. Warm-up training is also employed in this setting, followed by continued training with the LLM backbone.

\noindent \textbf{Language Structure Ablation.} 
We examine the impact of language diversity by restricting the number of plotting languages involved during training. For the language-controlled settings, we utilize strictly the target language scripts for each chart image. We further include an unaligned baseline where each chart image is paired with its original raw script from the source data described in Appendix \ref{sec:data_acquisition}.
To normalize the total training steps, we inversely scale the visual data: while the single-language models utilize the full 175K charts in the Chart2NCode training set, the two-language and three-language settings are restricted to random subsets of 1/2 (about 87.5K) and 1/3 (about 58.3K) of the chart images, respectively. The ensures a constant 175K chart-script pairs during training across all configurations. The training strategy is consistent with Section \ref{sec:training_strategy} for all configurations. We adjust the number of routers to match the target language count.

To validate that our findings are not artifacts of the downsampling strategy, we conduct a complementary ablation where the training budget is aligned to the three-language training without visual data reduction. In this regime, we allow all configurations to utilize the complete set of 175k unique chart images. To equate the computational cost with the three-language setting that trains on 525K chart-script pairs, we oversample the configurations with fewer targets: the single-language dataset is replicated three times, and the two-language dataset is augmented by randomly duplicating half of the available samples to reach the equivalent scale. Results in Figure~\ref{figure:ablation_router_language_appendix} show that the three-language model remains the top performer across all evaluation languages.

\noindent \textbf{Shared Subspace Ratio.} 
We visualize the subspace activation pattern of language-specific routers in ChartLuMA in Figure~\ref{figure:heatmap_subspace}. 
For quantification, we introduce the \textit{shared-subspace ratio}, which measures how much different language-specific routers rely on the same experts when processing the same chart.
Formally, for each chart example $c$, let $S_{c,l}\subseteq\{0,\dots,N-1\}$ denote the set of activated experts chosen by the router for language $l$, with $N=32$ in our standard setting. Each router activates a fixed number of experts (top--$k$, with $k=16$ in our experiments). Given the set of languages $\mathcal{L}_c$ available for chart $c$, we define $I_c = \bigcap_{l \in \mathcal{L}_c} S_{c,l}$ and $U_c = \bigcup_{l \in \mathcal{L}_c} S_{c,l}$, where $I_c$ is the set of experts shared across all languages and $U_c$ is the total set of experts activated by any language. The \textit{shared-subspace ratio} for chart $c$ is then $R_c = \frac{|I_c|}{|U_c|}$, which lies in $[0,1]$. A higher value indicates a dense shared core and a lower value implies strong language-specific specialization.

\subsection{Prompt Sensitivity Study}

To ensure robustness to specific lexical cues, we compare the standard ChartMimic prompt \citep{shi2024chartmimicevaluatinglmmscrossmodal} against two variants, while maintaining fixed system messages and output formats. The first variant strips contextual framing to retain only the core directive: \textit{``Generate the <language> code to reproduce the chart in this image.''} The second variant employs alternative wording: \textit{``Create a script in <language> that renders the figure shown. Ensure the output matches the visual details of the provided image.''} As shown in Table \ref{table:prompt_sensitivity}, the performance variance across these three settings is negligible, confirming that the model's capabilities are robust to instructional phrasing rather than being artifacts of a specific prompt template.

\begin{table} 
    \belowrulesep=0pt
    \aboverulesep=0pt
    \renewcommand{\arraystretch}{1.2}
    \setlength{\belowcaptionskip}{0.1cm}
    \centering
    \setlength\tabcolsep{3pt}
    \scalebox{0.8}{
    \begin{tabular}{p{2.7cm}<{\centering}|p{2.0cm}<{\centering}|p{1.2cm}<{\centering}p{1.2cm}<{\centering}p{1.2cm}<{\centering}}
        \toprule
         \multirow{2}{*}{\makecell[c]{Model}} & \multirow{2}{*}{\makecell{Prompt \\ Version}}  & \multicolumn{3}{c}{\textbf{Chart2NCode}} \\
        \cmidrule(r){3-5}
         & & ER & DS & MJ  \\
        \cmidrule(r){1-5}
        \multirow{3}{*}{Claude-Sonnet-4} & Default & 94.9 & 81.6  & 81.1   \\
        & Variant 1 & 94.9 & 81.7  & 81.2 \\
        & Variant 2 &  95.1 & 81.6  & 81.2 \\
        \midrule
        \multirow{3}{*}{Qwen3-VL-8B} & Default & 80.7 & 74.4  & 65.0   \\
        & Variant 1 & 80.8 & 74.3  & 64.8\\
        & Variant 2 &  80.5 & 74.3  & 65.1 \\
        \midrule
        \multirow{3}{*}{CharLuMA-1.3B} & Default & 91.1 & 78.9  & 72.3   \\
        & Variant 1 & 91.0 & 79.1  & 72.5 \\
        & Variant 2 & 91.2 & 79.0  & 72.3 \\
        \bottomrule
    \end{tabular}
    }
    \caption{Sensitivity study of evaluation prompt on the Chart2NCode test set.} 
    \label{table:prompt_sensitivity}
\end{table}

\subsection{Error Analysis}
\label{sec:errror_analysis}

\begin{figure*}[t]
    \centering
    \includegraphics[width=0.95\linewidth]{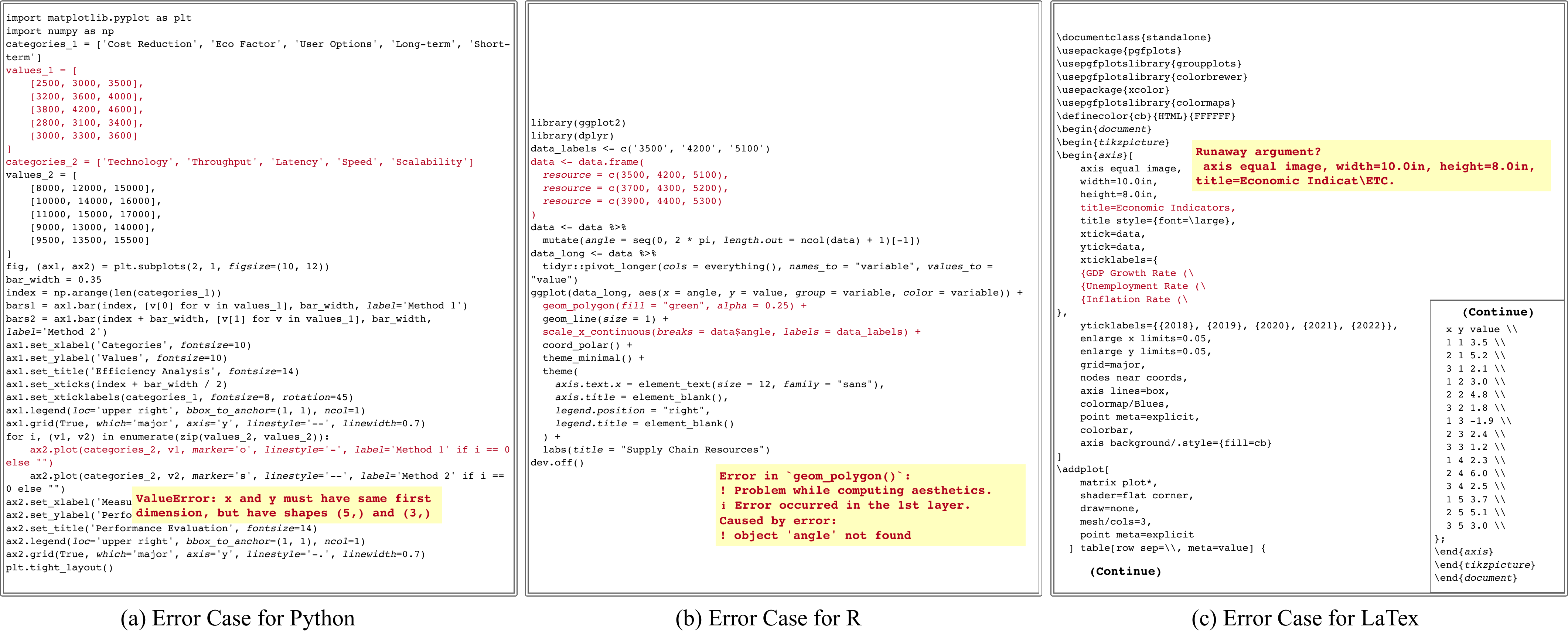}
    \caption{Case study of execution errors in generated code for CharLuMA-6.7B.}
    \label{fig:error_case_code}
\end{figure*}

We conduct an error analysis to identify the common sources of execution failures and reproduction limitations of CharLuMA-6.7B.
Execution failures in Python and R stem primarily from logic and data discrepancies, led by dimension mismatches (72.3\%, 56.1\%) and undefined variables (11.9\%, 22.0\%). In contrast, LaTeX errors are predominantly syntactic, with syntax omissions (55.5\%) significantly outweighing undefined variables (33.1\%) and dimension mismatches (11.4\%).
For example, the Python case in Figure~\ref{fig:error_case_code}(a) produces incompatible x–y list lengths when calling the \texttt{ax.plot} function. The R case in Figure~\ref{fig:error_case_code}(b) invokes an undefined variable \texttt{angle} in a \texttt{geom\_polygon} call. The LaTeX case in Figure~\ref{fig:error_case_code}(c) fails due to an omitted closing curly brace in the title and x-tick label definition.
Regarding reproduction fidelity, our qualitative assessment identifies annotation gaps as the dominant failure mode, manifested as mislabeled groups in Figure~\ref{fig:error_case_code}(a) or hallucinated text annotations in Figure~\ref{fig:error_case_code}(d). We also observe chart subtypes inaccuracies, illustrated by the generation of stacked instead of grouped error bars in Figure~\ref{fig:error_case_code}(b). Finally, stylistic inconsistencies remain prevalent, ranging from malformed x-ticks and incorrect ordering in R (Figure~\ref{fig:error_case_code}(c)) to deviant color schemes in LaTeX (Figure~\ref{fig:error_case_code}(d)).

\subsection{Comparison with Python Translation}

An intuitive approach to multi-language chart generation involves a two-step translation pipeline, wherein a model first generates a script in a primary language—such as Python—which is then translated into secondary formats like R or LaTeX. To benchmark this paradigm, we evaluate Qwen3-VL-8B and GPT-4o as chart-to-Python generators followed by a subsequent translation phase into R and LaTeX.
As evidenced in Table~\ref{table:comparison_translation}, this two-step process consistently compromises visual fidelity compared to direct generation. While using GPT-4o to translate the Python scripts of Qwen3-VL-8B can improve execution rates, it results in cascading errors that minor deviations in the initial Python code are amplified during translation, significantly degrading the final output's visual fidelity.
In contrast, CharLuMA-6.7B, trained on our Chart2NCode dataset, bypasses these intermediate bottlenecks by learning direct, universal visual-to-code mappings.
These findings demonstrate that a specialized, end-to-end approach is essential for achieving high-fidelity performance across diverse software ecosystems.

\begin{table}[t]
    \belowrulesep=0pt
    \aboverulesep=0pt
    \renewcommand{\arraystretch}{1.2}
    \setlength{\belowcaptionskip}{0.1cm}
    \centering
    \setlength\tabcolsep{3pt}
    \scalebox{0.72}{
    \begin{tabular}{p{2.4cm}<{\centering}p{2.4cm}<{\centering}|p{0.7cm}<{\centering}p{0.7cm}<{\centering}p{0.7cm}<{\centering}p{0.7cm}<{\centering}p{0.7cm}<{\centering}p{0.7cm}<{\centering}}
        \toprule
         \multirow{2}{*}{\makecell[c]{Generator}} & \multirow{2}{*}{\makecell{Translator}}  & \multicolumn{3}{c}{\textbf{Chart2R}} & \multicolumn{3}{c}{\textbf{Chart2LaTeX}} \\
        \cmidrule(r){3-5} \cmidrule(r){6-8}
         & & ER & DS & MJ & ER & DS & MJ  \\
        \cmidrule(r){1-8}
        \multicolumn{2}{c|}{\textit{Twp-step Translation}} \\
        Qwen3-VL-8B & Qwen3-VL-8B &  67.3          & 53.7          & 41.3          & 72.9          & 50.3          & 37.4          \\
        Qwen3-VL-8B & GPT-4o & 87.5          & 62.5          & 47.8          & 81.6          & 52.7          & 41.9          \\
        GPT-4o & Qwen3-VL-8B & 89.6          & 67.2          & 65.1          & 77.4          & 59.1          & 52.8          \\
        GPT-4o & GPT-4o &  95.3          & 73.6          & 71.3          & 82.4          & 64.2          & 60.7          \\
        \cmidrule(r){1-8}
        \multicolumn{2}{c|}{\textit{Direct Generation}} \\
        \multicolumn{2}{c|}{GPT-4o} & 94.5          & 78.8          & 78.3          & 88.4          & 72.4          & 69.8          \\
        \multicolumn{2}{c|}{Qwen3-VL-8B}  & 73.6          & 72.7          & 57.2          & 77.3          & 66.8          & 57.1          \\
        \multicolumn{2}{c|}{CharLuMA-1.3B}  & 94.5          & 78.9          & 73.3          & 84.5          & 71.3          & 65.1          \\
        \multicolumn{2}{c|}{CharLuMA-6.7B}  & \textbf{96.5} & \textbf{81.8} & \textbf{80.9} & \textbf{89.0} & \textbf{72.5} & \textbf{74.2} \\
        \bottomrule
    \end{tabular}
    }
    \caption{Performance comparison of direct chart-to-code generation and two-step translation from Python on the R and LaTeX subsets of Chart2NCode.} 
    \label{table:comparison_translation}
\end{table}

\subsection{Case Study}
\label{sec:examples}

\begin{figure*}[t]
    \centering
    \includegraphics[width=0.95\linewidth]{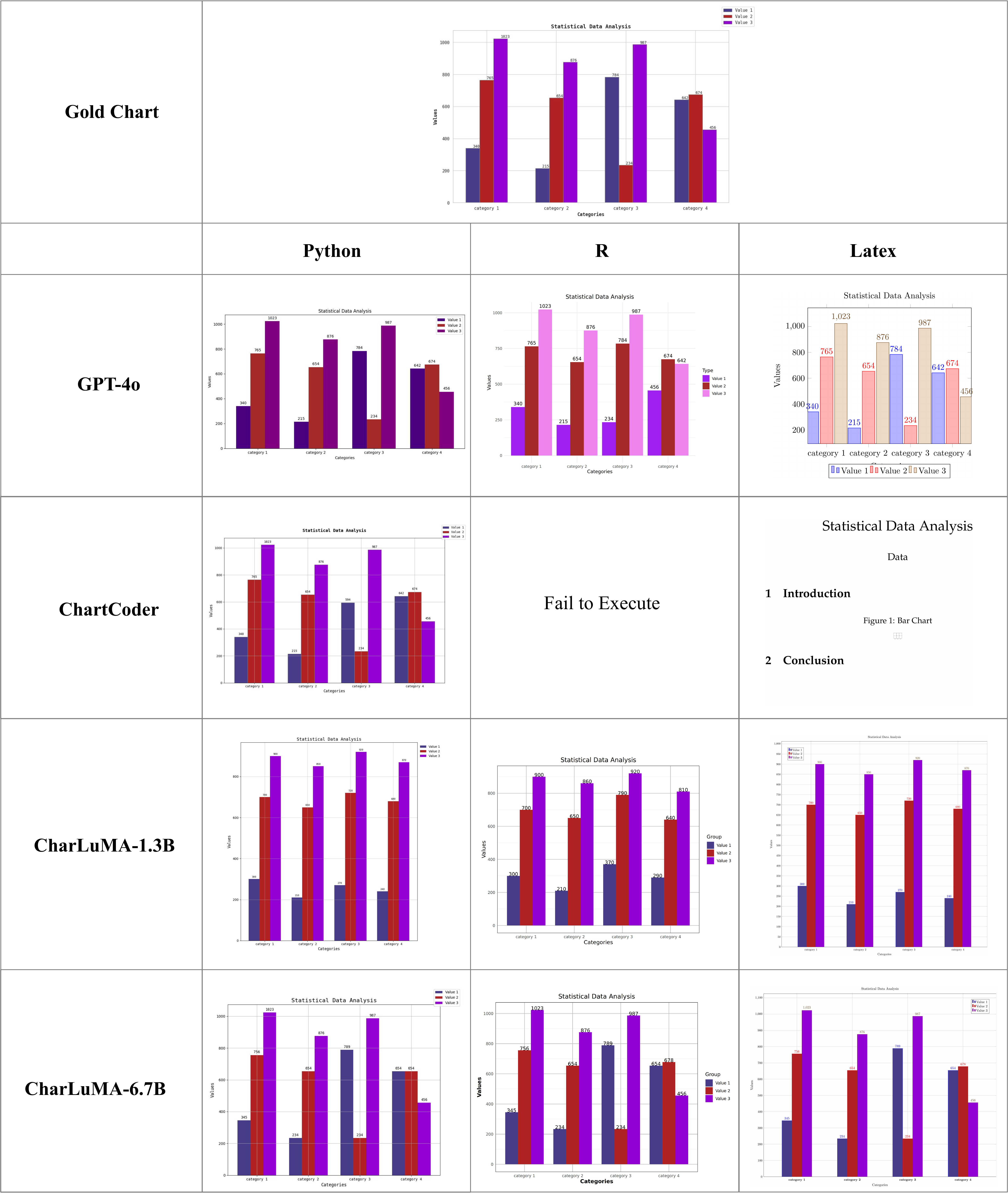}
    \caption{Case study of a grouped bar chart input and generated outputs from the Chart2NCode test set across three plotting languages.}
    \label{fig:model_example_1}
\end{figure*}

\begin{figure*}[t]
    \centering
    \includegraphics[width=0.95\linewidth]{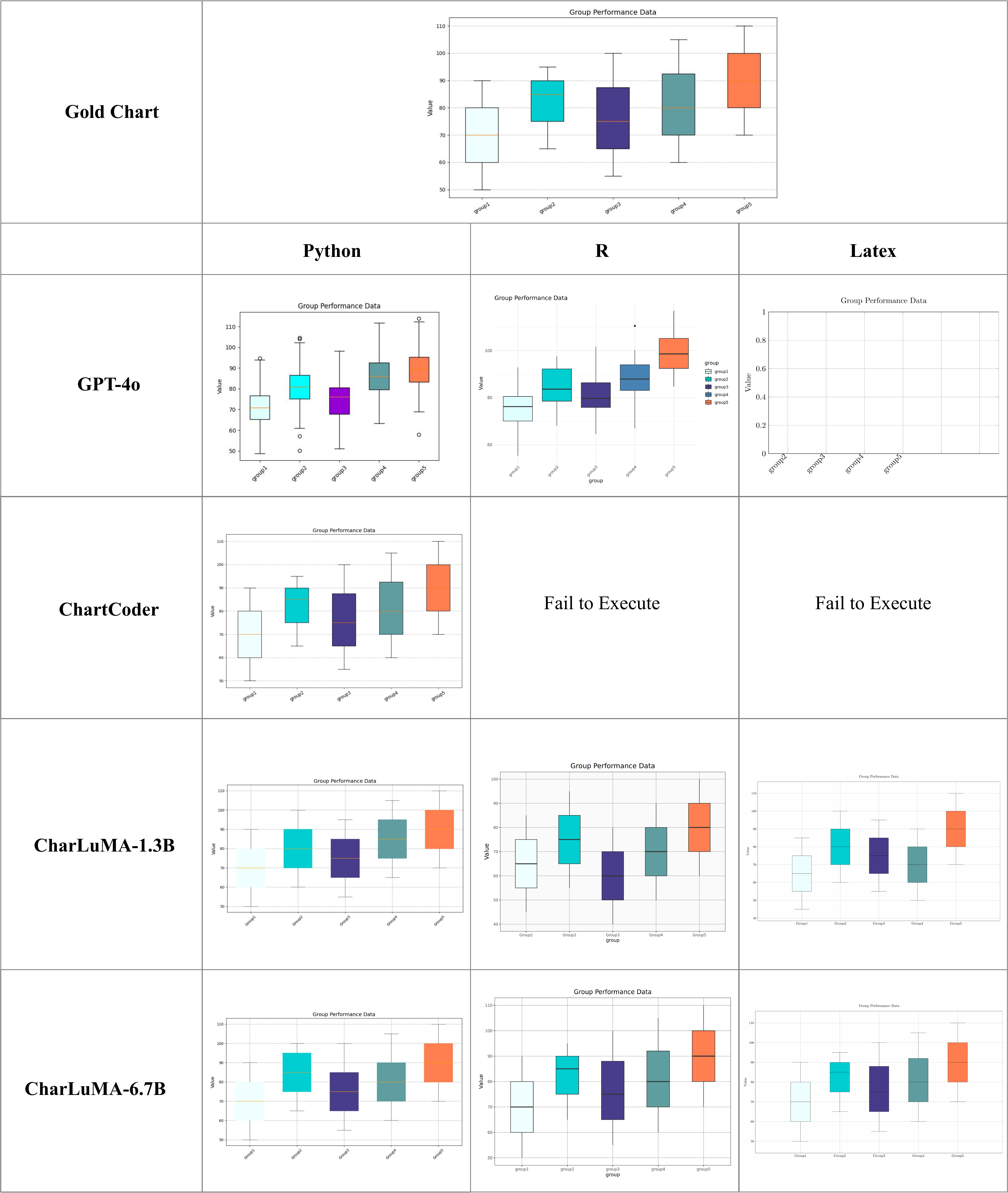}
    \caption{Case study of a box chart input and generated outputs from the Chart2NCode test set across three plotting languages.}
    \label{fig:model_example_2}
\end{figure*}

\begin{figure*}[t]
    \centering
    \includegraphics[width=0.95\linewidth]{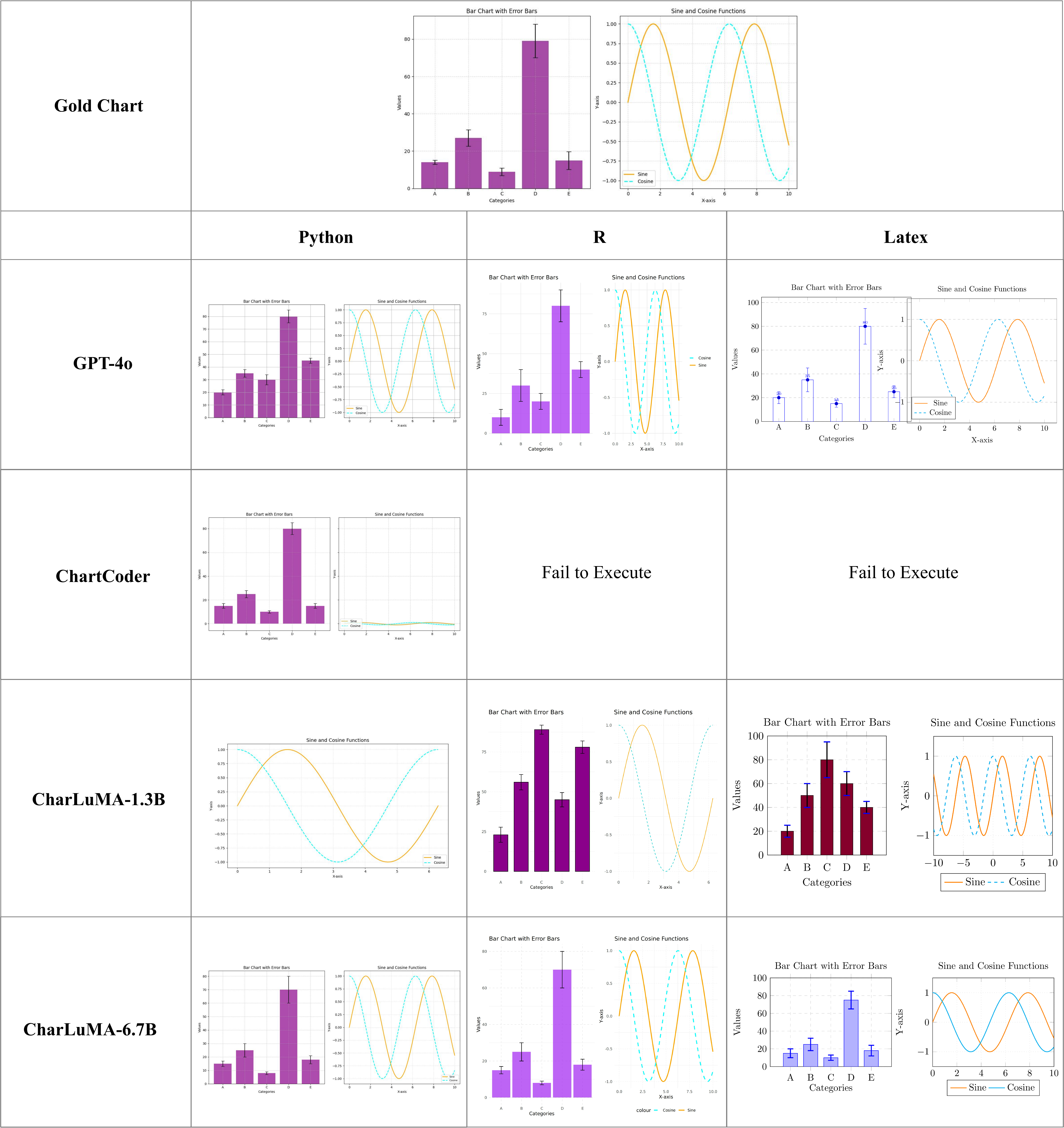}
    \caption{Case study of a two-subplot chart input and generated outputs from the Chart2NCode test set across three plotting languages.}
    \label{fig:model_example_3}
\end{figure*}

\begin{figure*}[t]
    \centering
    \includegraphics[width=\linewidth]{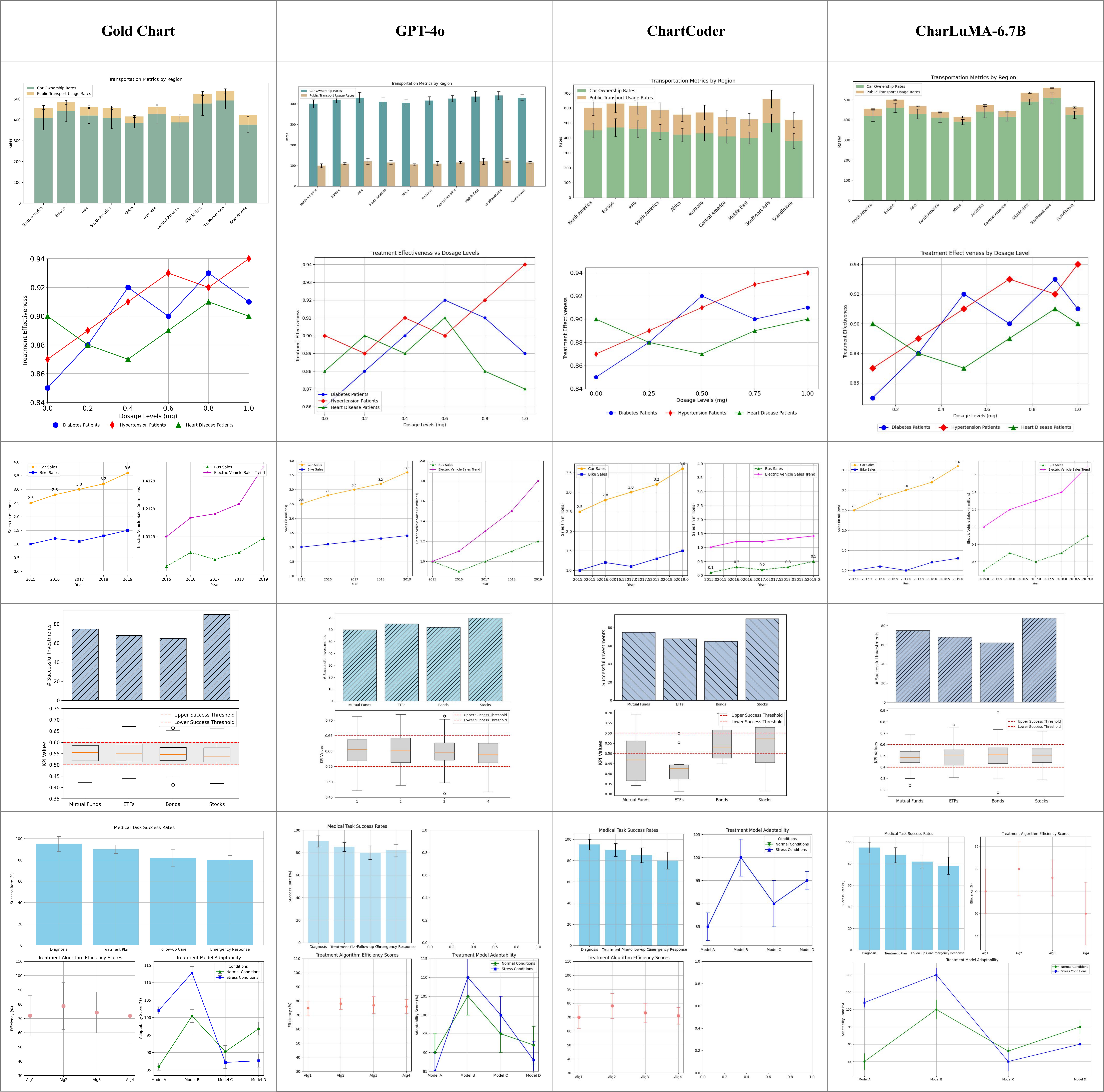}
    \caption{Case study of model inputs and generated outputs from ChartMimic in Python.}
    \label{fig:model_example_chartmimic}
\end{figure*}

We conduct a qualitative comparison of CharLuMA-6.7B against GPT-4o and ChartCoder using representative cases from Chart2NCode and ChartMimic.
In the Chart2NCode examples (Figure~\ref{fig:model_example_1}, Figure~\ref{fig:model_example_2}, and Figure~\ref{fig:model_example_3}), CharLuMA-6.7B demonstrates robust cross-language consistency, successfully reproducing high-quality charts where GPT-4o exhibits reduced reliability and ChartCoder frequently fails to generate valid R or LaTeX scripts.
Furthermore, through the four chart-to-Python examples from ChartMimic (Figure~\ref{fig:model_example_chartmimic}), we find that CharLuMA-6.7B matches the state-of-the-art performance of GPT-4o and ChartCoder, confirming its ability to handle advanced visual reasoning without compromising single-language proficiency.

\section{LLM Usage}

Large Language Models (LLMs) were used solely for grammatical and stylistic refinement of text originally drafted by the authors. They did not contribute to the research conceptualization, design, or analysis. The authors retain full responsibility for the accuracy and integrity of the final content.

\begin{figure*}[t]
\centering
\begin{tcolorbox}[
  title=Instruction Prompt for Handling Missing Templates in Post-Debugging,
  enhanced,
  sharp corners,
  halign=flush left,
  breakable=false
]
You are provided with a \texttt{\{original language\}} plotting script as shown below. Your task is to transform it to \texttt{\{target language\}} language, starting with ```\texttt{\{target language symbol\}} and ending with ```.

\texttt{\{original plotting script\}}
\end{tcolorbox}
\caption{Instruction prompt for handling missing templates in the post-debugging stage of the automatic annotation pipeline.}
\label{fig:prompt_debug_missing}
\end{figure*}

\begin{figure*}[t]
\centering
\begin{tcolorbox}[
  title=Instruction Prompt for Failed Template Execution in Post-Debugging,
  enhanced,
  sharp corners,
  halign=flush left,
  breakable=false
]

You are provided with two code snippets. The first is the original code, a \texttt{\{original language\}} plotting script serving as the reference implementation. The second is the transformed code, a version of the original script translated into \texttt{\{target language\}}, which is currently unexecutable due to syntax or logic errors.  

Original Code:
\texttt{\{original plotting script\}}

Transformed Code:
\texttt{\{failed template\}} 

Your task is to identify and correct all errors in the transformed code that prevent it from executing. The corrected script must produce a chart that is semantically equivalent to the one generated by the original code. High-level chart semantics such as axis labels, tick values, bar orientation, or grouping should remain unchanged unless modification is required for successful execution. You may reorder code lines, fix syntax issues, and adjust function arguments as needed. Please output only the corrected code, starting with ```\texttt{\{target language symbol\}} and ending with ```.  
\end{tcolorbox}
\caption{Instruction prompt for failed template execution in the post-debugging stage of the automatic annotation pipeline.}
\label{fig:prompt_debug_execut}
\end{figure*}

\begin{figure*}[t]
\centering
\begin{tcolorbox}[
  title=Prompt Template of Chart-to-code Generation Task,
  sharp corners,
  halign=flush left,
  breakable=false
]
You are an expert \texttt{\{target language\}} developer who specializes in writing code based on a given picture. I found a very nice picture in a STEM paper, but there is no corresponding source code available. I need your help to generate the \texttt{\{target language\}} code that can reproduce the picture based on the picture I provide. \\
Now, please give me the code that reproduces the picture below, starting with ```\texttt{\{target language symbol\}} and ending with ```.
\end{tcolorbox}
\caption{Prompt template of chart-to-code generation task (adapted from ChartMimic \citep{shi2024chartmimicevaluatinglmmscrossmodal}).}
\label{fig:prompt_chart2code}
\end{figure*}

\begin{figure*}[t]
\centering
\begin{tcolorbox}[
  title=Prompt Template of MLLM-as-Judge approach
  enhanced,
  sharp corners,
  halign=flush left,
  breakable=false
]
You are an excellent judge at evaluating visualization chart plots. The first image (reference image) is created using ground truth matplotlib code, and the second image (AI-generated image) is created using matplotlib code generated by an AI assistant. Your task is to score how well the AI-generated plot matches the ground truth plot.

\#\#\# Scoring Methodology:

The AI-generated image’s score is based on the following criteria, totaling a score out of 100 points:

1. Chart Types (20 points): Does the AI-generated image include all chart types present in the reference image (e.g., line charts, bar charts, etc.)?

2. Layout (10 points): Does the arrangement of subplots in the AI-generated image match the reference image (e.g., number of rows and columns)?

3. Text Content (20 points): Does the AI-generated image include all text from the reference image (e.g., titles, annotations, axis labels), excluding axis tick labels?

4. Data (20 points): How accurately do the data trends in the AI-generated image resemble those in the original image and is the number of data groups the same as in the reference image?

5. Style (20 points): Does the AI-generated image match the original in terms of colors (line colors, fill colors, etc.), marker types (point shapes, line styles, etc.), legends, grids, and other stylistic details?

6. Clarity (10 points): Is the AI-generated image clear and free of overlapping elements?

\#\#\# Evaluation:

Compare the two images head to head and provide a detailed assessment. Use the following format for your response:

— Comments:

- Chart Types: \$\{your comment and subscore\} 

- Layout: \$\{your comment and subscore\}

- Text Content: \${your comment and subscore} 

- Data: \$\{your comment and subscore\}

- Style: \$\{your comment and subscore\}

- Clarity: \$\{your comment and subscore\}

Score: \$\{your final score out of 100\} —

Please use the above format to ensure the evaluation is clear and comprehensive.
\end{tcolorbox}
\caption{MLLLM-as-Judge prompt template for chart-to-code generation evaluation (adapted from ChartMimic \citep{shi2024chartmimicevaluatinglmmscrossmodal})}
\label{fig:prompt_gpt_scoring}
\end{figure*}

\begin{figure*}[t]
    \centering
    \includegraphics[width=0.95\linewidth]{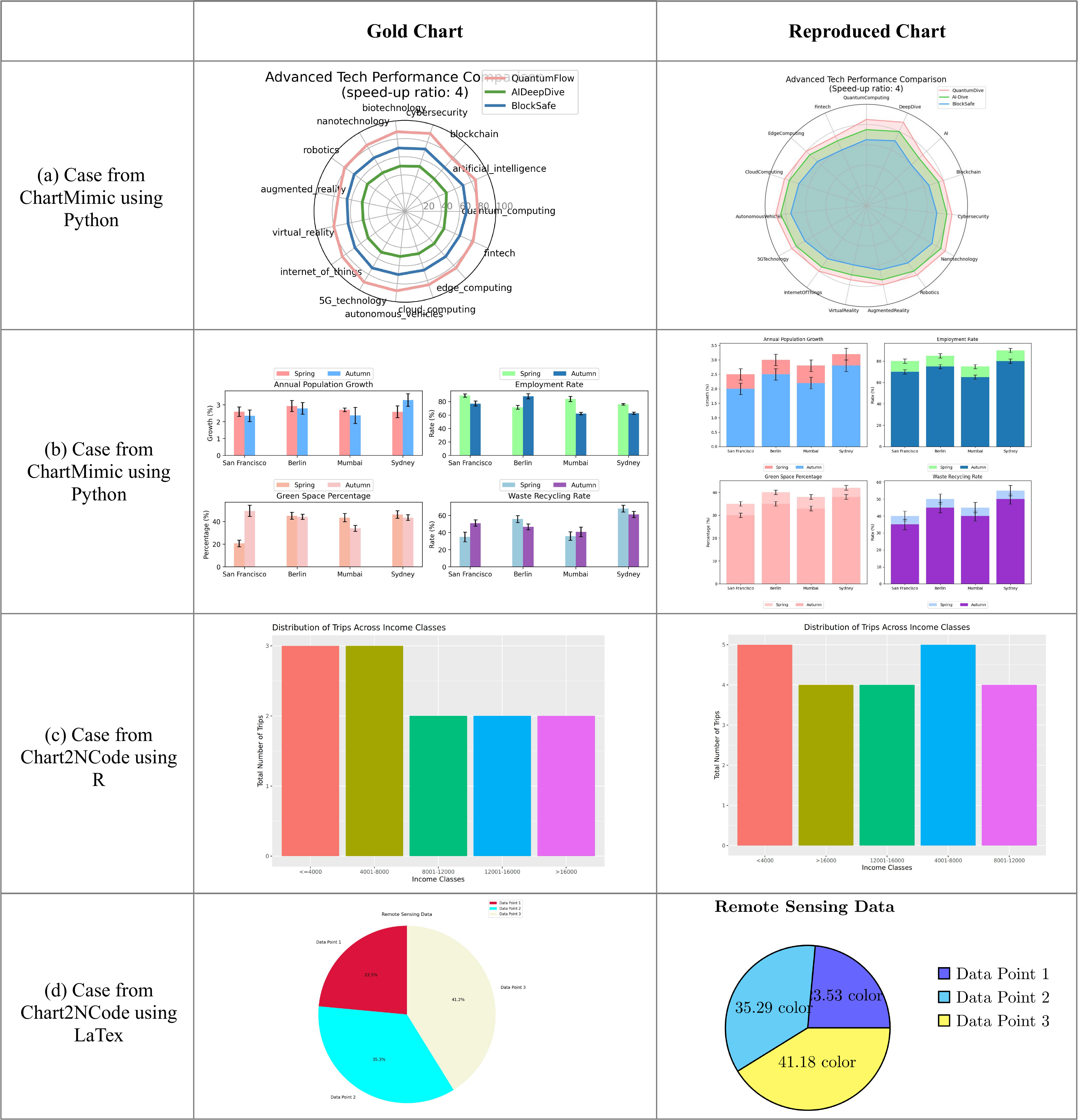}
    \caption{Case study of reproduction errors in generated charts for CharLuMA-6.7B.}
    \label{fig:error_case_image}
\end{figure*}

